\documentclass[Afour,sageh,times]{sagej}

\usepackage[utf8x]{inputenc}
\usepackage{moreverb,url}
\usepackage{glossaries}
\usepackage[caption=false]{subfig}
\usepackage{algorithm}
\usepackage{algpseudocode}
\usepackage{amsmath}
\usepackage{optidef}
\usepackage{gensymb}
\usepackage{adjustbox}
\usepackage[all]{nowidow}
\usepackage[hidelinks,bookmarksopen,bookmarksnumbered,citecolor=red,urlcolor=red]{hyperref}

\newcommand\BibTeX{{\rmfamily B\kercite-.05em \textsc{i\kern-.025em b}\kern-.08em
T\kern-.1667em\lower.7ex\hbox{E}\kern-.125emX}}

\def\volumeyear{2022}
\newcommand\palgmref[1]{\mbox{(Alg. \ref{alg:#1})}}
\newcommand\palgmlineref[2]{\mbox{(Alg. \ref{alg:#1}, #2)}}
\newcommand\talgmref[1]{\mbox{Algorithm \ref{alg:#1}}}
\newcommand\psecref[1]{\mbox{(Section \ref{sec:#1})}}
\newcommand\tsecref[1]{\mbox{Section \ref{sec:#1}}}
\newcommand\figref[1]{\mbox{(Fig. \ref{fig:#1})}}
\newcommand\tfigref[1]{\mbox{Fig. \ref{fig:#1}}}

\newcommand\ptblref[1]{\mbox{(Table \ref{tbl:#1})}}
\newcommand\ttblref[1]{\mbox{Table \ref{tbl:#1}}}

\newcommand\eqnlabel[1]{\label{eq:#1}}
\newcommand\algmlabel[1]{\label{alg:#1}}
\newcommand\seclabel[1]{\label{sec:#1}}
\newcommand\figlabel[1]{\label{fig:#1}}

\newcommand\tbllabel[1]{\label{tbl:#1}}

\newcommand\core{P_\mathrm{c}}
\newcommand\frontier{P_\mathrm{f}}
\newcommand\free{P_\mathrm{o}}

\newcommand\front{\mathbf{e}_\mathrm{f}}
\newcommand\bound{\mathbf{e}_\mathrm{b}}
\newcommand\viewf{\mathbf{v}}
\newcommand\frontp{\mathbf{f}}
\newcommand\viewp{\mathbf{x}}
\newcommand\viewo{\boldsymbol{\phi}}
\newcommand\view{\mathbf{v}\equiv(\viewp,\viewo)}
\newcommand\viewidx[1]{\mathbf{v}_#1\equiv(\viewp_#1,\viewo_#1)}
\newcommand\viewprime[1]{\mathbf{v}^#1\equiv(\viewp^#1,\viewo^#1)}

\newcommand\viewopt{\viewf^\star}

\newcommand\viewcurr{\viewf_\mathrm{c}}
\newcommand\viewoopt{\viewo^\star}
\newcommand\viewpopt{\viewp^\star}
\newcommand\viewoobs{\viewo_\mathrm{s}}

\newcommand\viewpcurr{\viewp_\mathrm{c}}
\newcommand\norm{\mathbf{e}_\mathrm{n}}
\newcommand\fvpair{\mathbf{m}\equiv(\frontp, \viewf)}

\newcommand\fvpairnmidx[1]{(\frontp_#1, \viewf_#1)}

\newcommand\fvpairm{\mathbf{m}}

\newcommand\frontd{\boldsymbol{s}}
\newcommand\frontdi{s}
\newcommand\point{\mathbf{p}}
\newcommand\qpoint{\mathbf{q}}
\newcommand\cpoint{\mathbf{c}}

\newcommand\npoint{\mathbf{n}}

\newcommand\cmass{\boldsymbol{\omega}}

\newcommand\algwhere{\;\textbf{where}\;}
\newcommand\algand{\;\textbf{and}\;}
\newcommand\algor{\;\textbf{or}\;}
\newcommand\Vprop{V}
\newcommand\Vobs{X_\mathrm{cap}}
\newcommand\Pall{P}
\newcommand\Pnew{P^\prime}
\newcommand\visited{R}
\newcommand\surfgeom{E_\mathrm{surf}}
\newcommand\Mall{S}
\newcommand\Mobs{S_\mathrm{reg}}
\newcommand\algnull{\textsc{null}}

\DeclareMathOperator*{\argmax}{arg\,max}
\DeclareMathOperator*{\argmin}{arg\,min}

\algnewcommand{\IfThen}[2]{\State \algorithmicif #1 \algorithmicthen #2}

\makeatletter
\newcommand{\mathleft}{\@fleqntrue\@mathmargin0pt}
\newcommand{\mathcenter}{\@fleqnfalse}
\makeatother

\begin{document}
	
\newacronym{ori}{ORI}{Oxford Robotics Institute}
\newacronym{drs}{DRS}{Dynamic Robot Systems}
\newacronym{inf}{IG}{Information Gain}
\newacronym{nbv}{NBV}{Next Best View}
\newacronym{mvs}{MVS}{Multi View Stereo}
\newacronym{see}{SEE}{Surface Edge Explorer}
\newacronym{prm}{PRM}{Probabilistic Roadmap}
\newacronym{cnn}{CNN}{Convolutional Neural Network}
\newacronym{gng}{GNG}{Growing Neural Gas}
\newacronym{pca}{PCA}{Principal Component Analysis}
\newacronym{pga}{PGA}{Principal Geodesic Analysis}
\newacronym{pns}{PNS}{Principal Nested Spheres}
\newacronym{pngs}{PNGS}{Principal Nested Great Spheres}
\newacronym{hpr}{HPR}{Hidden Point Removal}
\newacronym{slsqp}{SLSQP}{Sequential Least-Squares Quadratic Programming}
\newacronym{gp}{GP}{Gaussian Process}
\newacronym{ros}{ROS}{Robot Operating System}
\newacronym{dbscan}{DBSCAN}{Density-Based Spatial Clustering of Applications with Noise}
\newacronym{icp}{ICP}{Iterative Closest Point}
\newacronym{oumnh}{OUMNH}{Oxford University Museum of Natural History}
\newacronym{tsp}{TSP}{Travelling Salesman Problem}
\newacronym{ugv}{UGV}{Unmanned Ground Vehicle}
\newacronym{uav}{UAV}{Unmanned Aerial Vehicle}
\newacronym{auv}{AUV}{Autonomous Underwater Vehicle}
\newacronym{rrt}{RRT}{Rapidly-exploring Random Trees}
\newacronym{rrg}{RRG}{Rapidly-exploring Random Graph}
\newacronym{mcmc}{MCMC}{Markov Chain Monte Carlo}
\newacronym{tsdf}{TSDF}{Truncated Signed Distance Field}
\newacronym{ompl}{OMPL}{Open Motion Planning Library}
\newacronym{aitstar}{AIT*}{Adaptively Informed Trees}
\newacronym{knn}{\textit{k}-NN}{\textit{k}-Nearest Neighbours}
\newacronym{nerf}{NeRF}{Neural Radiance Field}

\runninghead{Border and Gammell}

\title{The Surface Edge Explorer (SEE): A measurement-direct approach to next best view planning}

\author{Rowan Border\affilnum{1} and Jonathan D. Gammell\affilnum{1}}

\affiliation{\affilnum{1}Estimation, Search, and Planning (ESP) Research Group, Oxford Robotics Institute (ORI), Department of Engineering Science, University of Oxford, Oxford, United Kingdom}

\corrauth{Rowan Border,
Oxford Robotics Institute,
23 Banbury Road,
Oxford,
United Kingdom,
OX2 6NN}

\email{rborder@oxfordrobotics.institute}

\begin{abstract}

High-quality observations of the real world are crucial for a variety of applications, including producing 3D printed replicas of small-scale scenes and conducting inspections of large-scale infrastructure. These 3D observations are commonly obtained by combining multiple sensor measurements from different views. Guiding the selection of suitable views is known as the \gls{nbv} planning problem.

Most \gls{nbv} approaches reason about measurements using rigid data structures (e.g., surface meshes or voxel grids). This simplifies next best view selection but can be computationally expensive, reduces real-world fidelity, and couples the selection of a next best view with the final data processing.

This paper presents the \gls{see}, a NBV approach that selects new observations directly from previous sensor measurements without requiring rigid data structures. \gls{see} uses measurement density to propose next best views that increase coverage of insufficiently observed surfaces while avoiding potential occlusions. Statistical results from simulated experiments show that \gls{see} can attain similar or better surface coverage with less observation time and travel distance than evaluated volumetric approaches on both small- and large-scale scenes. Real-world experiments demonstrate \gls{see} autonomously observing a deer statue using a 3D sensor affixed to a robotic arm.
       
\end{abstract}

\keywords{3D reconstruction, active vision, view planning, next best view, pointcloud representation, measurement-direct approach}

\maketitle

\glsresetall

\section{Introduction}

Capturing high-fidelity observations of the real world is crucial for performing accurate analysis. High-accuracy scanners attached to industrial robots can be used to compare the structure of manufactured parts with ground-truth production models for quality control. Observations obtained from surveying large-scale outdoor structures, typically with an aerial platform, can be used for infrastructure inspection or to preserve edifices of historical significance. For example, observations of the Notre-Dame de Paris and the ancient city of Palmyra are being used to aid their respective reconstruction efforts.

Obtaining high-quality 3D observations is a challenge regardless of their final purpose. A scene (i.e., a bounded region of space) is observed by combining individual 3D measurements of surfaces obtained from multiple different positions. An observation is complete when there is sufficient measurement coverage on all visible surfaces. The final surface coverage achieved depends on the sensor capabilities, the scene structure, and the views from which measurements are obtained. These views can be chosen by a human operator, but empirically selecting views is often undesirable or impossible.

Algorithmic view selection mitigates human uncertainty by intelligently choosing views. This challenge of planning a \emph{next} view that can provide the \emph{best} improvement in a scene observation is known as the \gls{nbv} planning problem. It was first explored by \citet{Connolly1985}.

\gls{nbv} approaches can be broadly categorised by their sources of information. Model-based approaches require prior scene information to plan views (e.g., to compare a manufactured part with its production model) and cannot generalise to unknown scenes. Model-free approaches do not require \emph{a priori} scene information and plan next best views from the current observation state.

The state of an observation is encoded with a scene representation. Most model-free \gls{nbv} planning approaches use \emph{structured} representations. These impose an external structure onto the scene. Volumetric representations segment the scene volume into a 3D voxel grid. Surface representations create a connected mesh from subsampled sensor measurements. These representations aggregate multiple measurements into each element of their structure. This simplifies selecting next best views but is computationally expensive and reduces observation fidelity (e.g., measurements from a partial surface may be sufficient for a voxel to be considered observed or a mesh to be connected without preserving surface details). Increasing the structural resolution (e.g., voxel number or mesh density) captures more detail but also increases computational costs. High resolution voxel grids and denser meshes are more computationally expensive to raycast and update.

\begin{figure}[tpb]
	\centering
	\includegraphics[width=0.65\linewidth]{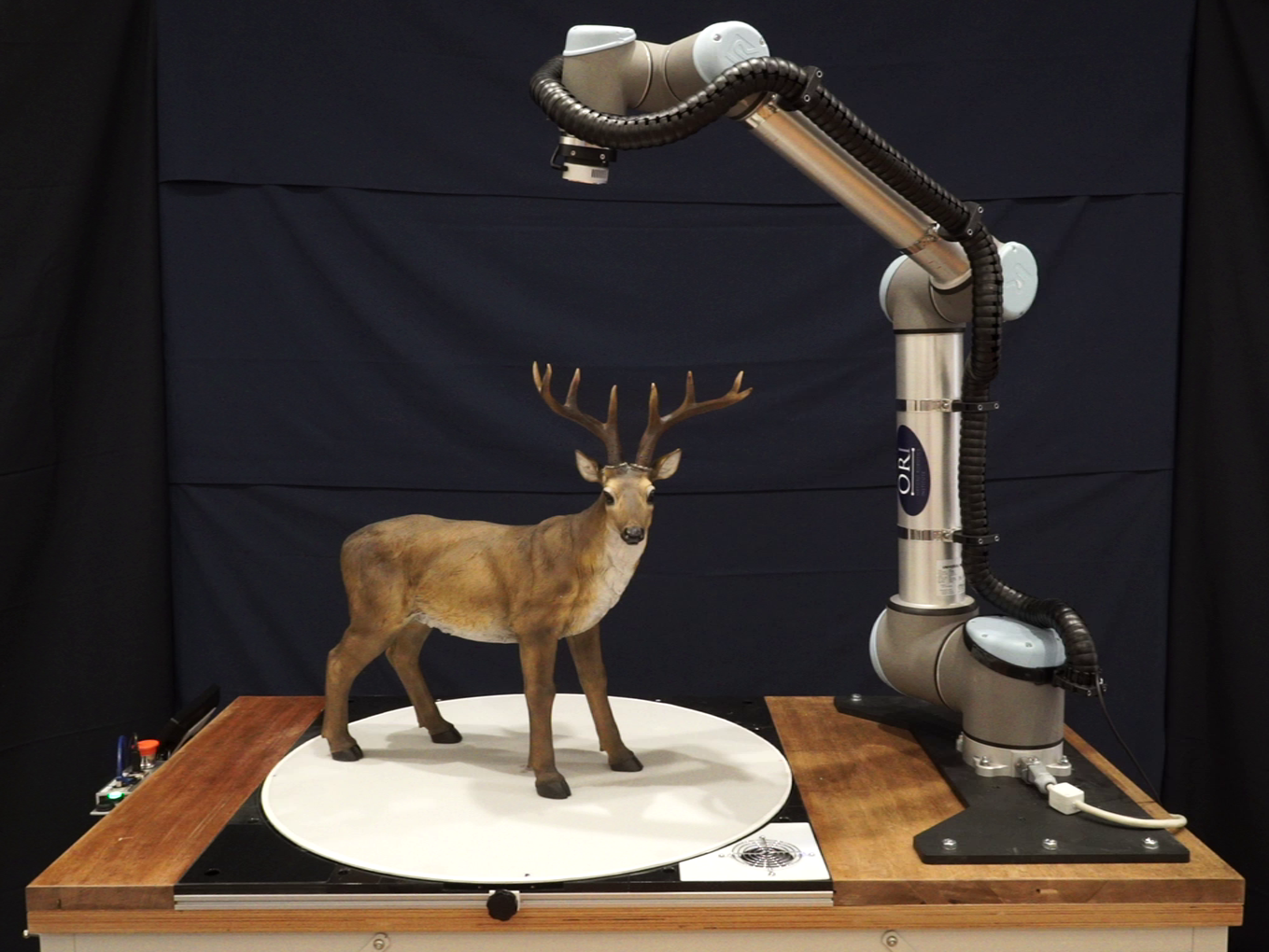}
	\caption{A photograph of the UR10 platform. An object (e.g., the Oxford Deer) is placed on a turntable (white) and captured by an Intel RealSense L515 on the UR10 end effector. SEE jointly directs the turntable and UR10 to observe the object.}
	\figlabel{inspector-real}
\end{figure} 

\gls{nbv} approaches with unstructured pointcloud representations do not impose external structures onto the scene and reason directly about sensor measurements. This maintains full fidelity and does not require restrictive assumptions about the scene structure but does require NBV approaches to reason about sensor measurements. Approaches with pointcloud representations aim to obtain coverage of scene surfaces rather than volumes of space by capturing new measurements that incrementally expanded an observation over connected surfaces.

The \gls{see} proposes next best views directly from measurement density and is therefore referred to as a `measurement-direct' approach. It identifies surfaces with low measurement density and proposes views to capture additional measurements. These views can be refined to avoid occlusions and improve visibility of their target surfaces. A next best view is chosen from these proposals to obtain a significant improvement in scene coverage while moving a short distance. New measurements from this view are then added to the observation and new views are incrementally captured until a minimum measurement density is obtained from the entire scene. The efficiency of this method allows SEE to obtain complete observations with short travel distances and observation times compared to state-of-the-art volumetric approaches.

The observation performance of \gls{see} is evaluated with both simulated and real-world experiments. The simulation experiments provide a quantitative comparison of \gls{see} with state-of-the-art volumetric approaches on both small- and large-scale scenes. In one simulation, small-scale tabletop models are observed using an RGB-D camera attached to a robotic arm. In another simulation, large-scale building models are observed with a LiDAR mounted on an aerial platform. \gls{see} consistently outperforms the evaluated volumetric approaches in these simulation experiments by obtaining similar or better surface coverage while travelling shorter distances and requiring less observation time.

\begin{figure}[tpb]
	\centering
	\includegraphics[width=0.5\linewidth]{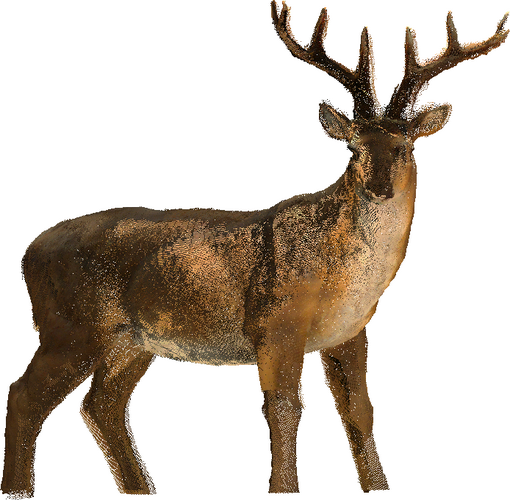}
	\caption{A pointcloud of the Oxford Deer captured by SEE using the UR10 platform.}
	\figlabel{marquee}
\end{figure}      

Real-world experiments demonstrate the same observation performance on a physical robotic platform \figref{inspector-real}. The platform is comprised of a turntable, onto which observation targets are placed, and a UR10 robot arm with an Intel RealSense L515 affixed to the end effector. It was specifically designed to evaluate \gls{nbv} approaches and is not intended to be comparable with commercial 3D scanners. The observation target used for the real-world experiments was a deer statue, known as the Oxford Deer. Results from these real-world experiments show that \gls{see} can obtain high-quality observations of the Oxford Deer with similar performance to the simulations, despite sensor noise \figref{marquee}. 

\subsection{Statement of Contributions}

This paper presents a definitive version of \gls{see} and evaluates its observation capabilities in simulation and the real world. \gls{see} and its density-based pointcloud representation were first presented in \citet{Border2017} and extended in \citet{Border2018}, \citet{BorderThesis} and \citet{Border2019}. This paper makes the following specific contributions:  

\begin{itemize}
	\item Presents a definitive version of \gls{see} that unifies previously published work and extends it to improve performance. These extensions include computing suitable parameters online instead of using user-specified values and a more robust method for determining the correct direction of surface normals. 
	\item Uses realistic simulations of both small- and large-scale scenes to compare the observation performance of SEE with state-of-the-art volumetric approaches. The results demonstrate that \gls{see} can capture highly complete scene observations using less travel distance and observation time than the volumetric approaches.
	\item Presents the first evaluation of SEE working with a fully autonomous real-world system. This illustrates that \gls{see} can efficiently obtain high-quality observations of a real-world object using a 3D sensor.    
\end{itemize}

\begin{figure*}[tpb]
	\centering
	\captionsetup{justification=centering}
	\subfloat[Volumetric representation]{\includegraphics[width=0.3\linewidth]{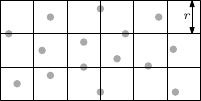}} \quad
	\subfloat[Surface representation]{\includegraphics[width=0.275\linewidth]{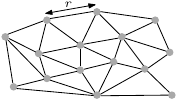}} \quad
	\subfloat[Pointcloud representation]{\includegraphics[width=0.275\linewidth]{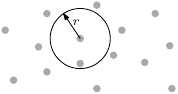}} 
	\captionsetup{justification=justified}
	\caption{Illustrations of (a) a volumetric scene representation (i.e., a voxel grid), (b) a surface representation (i.e., a connected mesh) and (c) the density-based pointcloud representation presented in this paper. A resolution parameter, $r$, defines the voxel size for a volumetric representation, the maximum edge length for a surface representation or the search radius for the density representation.}
	\figlabel{back-rep}
\end{figure*}  

\section{Related Work}
\seclabel{back}

This section presents an overview of the \gls{nbv} planning problem \psecref{nbv-prob} and a review of relevant literature on model-free approaches. Different strategies for categorising \gls{nbv} planning approaches have been presented in survey papers \citep{Tarabanis1995, Scott2003a, Karaszewski2016a, Zeng2020a}. This paper adopts the classification scheme used by \citet{Scott2003a} to discuss approaches based on their scene representation. 

Approaches using volumetric representations are reviewed in \tsecref{vol-rep}, those with surface representations are discussed in \tsecref{surf-rep} and approaches that utilize a combination of multiple representations are considered in \tsecref{comb-rep}. A new class of pointcloud representations, which includes the density representation used by \gls{see}, is introduced in \tsecref{free-rep}.

\subsection{The Next Best View Planning Problem}
\seclabel{nbv-prob}

The challenge of \gls{nbv} planning is proposing and selecting views from which a scene can be efficiently observed. The quality of an observation can be quantified by its accuracy (i.e, how closely the captured data resembles the actual scene) and completeness (i.e., what proportion of the scene has been observed). Observation accuracy primarily depends on the sensor capabilities but can be improved by considering the scene texture and geometry. The completeness of an observation is determined by the coverage obtained from captured views. The efficiency of a \gls{nbv} approach is quantified by the time and travel distance required to obtain an observation.

Approaches to the \gls{nbv} planning problem select views by evaluating information obtained from previous views and in some cases \emph{a priori} scene information. The general \gls{nbv} problem can be formally expressed as a function, $\viewf^\prime = \mathrm{nbv}(V,Y)$, which selects a view, $\viewf^\prime$, from a (possibly uncountable) set of potential views, $V$, based on information obtained from previous views or \emph{a priori} scene information, $Y$. A next best view is selected to provide the greatest observation improvement by evaluating a set of quantitative metrics, $\mathrm{metrics}(\viewf, Y)$,

\begin{equation*}
\eqnlabel{nbv-prob}
\viewf^\prime = \mathrm{nbv}(V, Y) = \argmax_{\viewf \in V}\; \mathrm{metrics}(\viewf, Y)\,.   \
\end{equation*}

Each captured view improves the scene information available for selecting subsequent next best views. The representation chosen to encode this information is a defining characteristic of \gls{nbv} algorithms. Volumetric representations segment the scene volume into a three-dimensional voxel grid (\tfigref{back-rep}a). Surface representations connect sensor measurements to create a surface mesh (\tfigref{back-rep}b). Some \gls{nbv} planning approaches also encode scene information using a combination of these volumetric and surface representations. Pointcloud representations, such as the density representation presented in this paper, do not impose an external structure on the scene or assume any connectivity between sensor measurements (\tfigref{back-rep}c).

The \gls{nbv} selection metrics used by an approach are largely determined by its representation. Approaches with volumetric representations typically aim to achieve the greatest reduction in the volume of unobserved space. Approaches with surface representations usually aim to expand the extent of their mesh and reduce surface uncertainty. Approaches with pointcloud representations commonly aim to improve the coverage and density of captured measurements. Almost every approach also considers the travel cost incurred to reach a view.    
    
\subsection{Volumetric Representations}
\seclabel{vol-rep}

\gls{nbv} planning approaches most commonly use a \emph{volumetric} representation for encoding scene information. This representation divides the scene volume into three-dimensional cells known as voxels. A state associated with each voxel encodes information on its observation status and occupancy. The visibility of voxels from a set of potential views is evaluated by raycasting the voxel grid from each view and recording the states of voxels intersected by the rays. A value for each view is determined from the intersected voxel states.




Approaches using a volumetric representation can be broadly separated into three categories based on their method for proposing views. Many approaches use a predefined set of views chosen from a surface (e.g., a sphere or hemisphere) encompassing the scene \psecref{vol-rep-pre}. Some approaches use path planning methods to propose views within free space regions of the scene \psecref{vol-rep-plan}. Other approaches propose views using scene information obtained from  the occupancy and observation states of voxels \psecref{vol-rep-inf}.

\subsubsection{Selecting Views From a Predefined Set}
\seclabel{vol-rep-pre}

Volumetric approaches that select views from a predefined set are primarily differentiated by how they use the measurement information encoded in their voxel representation.

\citet{Connolly1985} coined the term \emph{next best view} in formative work which presents the first approaches to the \gls{nbv} problem. Voxels are classified by their occupancy and observation states. Next best views are selected to capture the most unobserved voxels. \citet{Papadopoulos-Orfanos1997a,Wong1999} use the same voxel classifications and view selection metric but different view proposal sets (e.g., using the centers of empty voxels). These approaches prioritise capturing unobserved voxels and consider occlusions from occupied voxels but do not classify occluded voxels or evaluate observation quality.   

Several approaches extend this work by classifying occluded voxels and considering observation quality. \citet{Banta2000} introduce an occluded classification for voxels that lie within a viewing frustum but are obscured by occupied voxels. \citet{Massios1998,Vasquez-Gomez2009,Vasquez-Gomez2014} present approaches that combine an occluded voxel classification with view selection metrics that consider observation quality. These approaches obtain better measurements by explicitly considering occlusions and observation quality but the use of binary voxel occupancy states can result in scene regions being sparsely covered by captured measurements. 

Some limitations of binary voxel states can be addressed with a probabilistic voxel representation. This defines voxel occupancy as the likelihood that a voxel should contain sensor measurements. NBV approaches using this representation select next best views with \gls{inf} metrics, which quantify view value as the expected information available from voxels. \citet{Potthast2014,Isler2016,Delmerico2017} present \gls{inf}-based view selection metrics that consider the occupancy probability, visibility and distance of voxels from a view. \citet{Abduldayem2017,Almadhoun2019} use knowledge of the scene geometry to predict voxel occupancy. 

\citet{Curless2011} create an implicit \gls{tsdf} surface \citep{Curless1996} and select views to reduce its uncertainty. \citet{Hou2019} improve upon the accuracy of independent per-voxel occupancy probabilities by jointly estimating occupancy probabilities for the entire voxel grid. \citet{Lauri,Lauri2020} jointly maximise an \gls{inf} metric between multiple sensors while reducing unnecessary overlap between their views. These probabilistic approaches are able to obtain observations with more consistent scene coverage by encoding more detailed measurement information in voxels, but are still limited by the voxel grid resolution.  

Some recent works \citep{Wang2019,Mendoza2020,Pan2022,Pan2023a} have applied learning methods to the \gls{nbv} problem by training networks to select next best views from a predefined set. These approaches can obtain highly complete observations of scenes with geometry similar to the training sets but do not necessarily generalise to scenes with unseen geometry.  

Many volumetric approaches that choose views from a predefined set are capable of obtaining observations with high coverage of the scene volume; however, the completeness of these observations depends on the distribution of the predefined views since high coverage of the scene volume does not ensure good surface visibility. 

\subsubsection{Sampling Views With Path Planning}
\seclabel{vol-rep-plan}

Other volumetric approaches use sampling-based planning techniques to generate views in the free space of a scene. Views are proposed at the sampled states and evaluated when selecting a next best view.

Some approaches sample views from the entire scene volume. This enables them to obtain high global scene coverage but it is computationally expensive to raycast a large number of views.  \citet{Potthast2011} use the \acrlong{prm}  \citep[\acrshort{prm};][]{Kavraki1996}\glsunset{prm} planner. \citet{Yoder2016} use the SPARTAN planner \citep{Cover2013}. 

Many approaches \citep{Vasquez-Gomez2015,Vasquez-Gomez2018,Bircher2016,Bircher2018a,Respall2021a} reduce the cost of raycasting views by limiting their view sampling to a local region around the current sensor position \citep[e.g., using \gls{rrt} or RRT*;][]{LaValle1998,Karaman2011}. Observations captured by these approaches can be highly complete in local scene regions but do not typically obtain good global coverage.     

Recent approaches introduce methods to obtain high global scene coverage with a reduced computational cost. \citet{Selin2019} extend \citet{Bircher2016,Bircher2018a} by preserving high-value views between sampling iterations and evaluating view value with an efficient sparse raycasting method. \citet{Schmid2019} use \gls{rrt}* to expand a single exploration tree when sampling new views and use a \gls{tsdf} map representation to consider observation quality when selecting views. \citet{Schmid2022a} extend this work by training a neural network to learn a voxel confidence metric for selecting views. \cite{Dang2019} build both local and global exploration graphs by sampling views with a \gls{rrg} algorithm \citep{Karaman2009}. \cite{Xu2021a} propose an incremental \gls{prm}-based algorithm for sampling views. \cite{Cao2023} train a reinforcement learning network to select a next best view from a set of views uniformly sampled in free space. 

Volumetric approaches that sample view proposals using path planning techniques can typically obtain greater scene coverage than those using a predefined set of views. The sampled views have better visibility of scene surfaces since they are distributed throughout the scene volume instead of around it. A limitation is the computational cost of raycasting a large number of sampled views, which is an important consideration when choosing a feasible sampling density.  

\subsubsection{Proposing Views Using Scene Information}
\seclabel{vol-rep-inf}

Volumetric approaches that use the current voxel states to propose views can capture high-quality observations with greater efficiency than other volumetric methods as higher quality views are proposed and fewer views need to be considered.

Some approaches simply propose views in unoccupied voxels \citep{Potthast2014, Daudelin2017, Palomeras2019}. These are unlikely to obtain higher quality observations as they do not consider the observed scene geometry when proposing views but may achieve lower computation times by evaluating fewer views. 

The highest quality observations are captured when views are proposed using the observed scene geometry. Volumetric approaches can achieve this by proposing views to capture surface frontier voxels (i.e., occupied voxels with unobserved neighbours) or exploration frontier voxels (i.e., unoccupied voxels with unobserved neighbours). Some approaches \citep{Monica2018,Hardouin2020,Hardouin2020a} identify clusters of surface frontier voxels, compute a normal for each cluster and generate a view along each normal. \citet{Kompis2021b} compute a normal for each surface frontier voxel and generate multiple views around each normal. \citet{Batinovic2021a} use a mean-shift clustering approach to identify views that lie at the center of exploration frontier clusters. These approaches typically select a next best view to capture the most frontier voxels while moving the least distance.

Some recent approaches \citep{Schmid2022,Zacchini2023,Ren2023} use learning methods to generate views based on scene information. They learn to propose views in free space that are likely to improve coverage of the scene by training on views generated by an existing sampling algorithm \citep{Schmid2022, Ren2023} or by learning a scene-specific view distribution online from the current observation state \citep{Zacchini2023,Ren2023}. These approaches demonstrate promising results in comparison with traditional volumetric methods for proposing views.            

Volumetric approaches that use the current voxel states to propose views can obtain higher quality observations with greater efficiency than other volumetric methods. Computational cost is reduced as only views proposed to improve an observation are evaluated with raycasting.

\subsection{Surface Representations}
\seclabel{surf-rep}

Approaches with \emph{surface} representations approximate the scene geometry by creating a connected mesh from sensor measurements. This mesh can provide high-fidelity information about the scene structure for proposing views, selecting a next best view and evaluating observation quality.

Many surface approaches require multiple data capture stages to obtain complete scene observations. An initial observation creates a course mesh from sparse measurements by following a preplanned or human-directed view trajectory. This initial mesh is then refined by integrating measurements from a second \gls{nbv}-planning-directed observation.

\citet{Reed2000} select next best views to observe occluded surfaces in a mesh representation. This improves the measurement density on surfaces with irregular geometry and refines the initial mesh observation. \citet{Hollinger2012} model uncertainty in the initial mesh using Gaussian process implicit surfaces and select next best views to reduce the uncertainty. \citet{Roberts2017} propose views by independently sampling sets of view positions and orientations. The value of each view is defined by the visibility of vertices in the initial mesh and a view trajectory is planned to maximise the additive value of each view visited. \citet{Peng2018a} account for the scene geometry when proposing views by sampling them from a manifold encompassing the scene, which is produced by moving each vertex in the initial mesh a given distance along its surface normal and computing a convex hull. 

All these two-stage approaches are able to capture high quality observations by using knowledge of the scene geometry obtained from the initial survey, but requiring an extra capture stage increases the overall observation time. 

Some surface-based approaches do not require a multistage observation. \citet{Khalfaoui2013} classifies the visibility of surfaces in its current mesh based on the angles between their normals and the sensor pose. Views are chosen to observe surfaces with poor visibility. \citet{Lim2023a} present a \gls{nbv} approach for \gls{mvs} reconstruction that selects views from a predefined set to provide the best coverage of surface landmarks.

Surface approaches can typically capture higher quality scene observations than volumetric approaches by considering the scene geometry when proposing and selecting views; however, requiring multiple observation stages increases the overall capture time and creating a mesh from dense sensor measurements is often computationally expensive. 

\subsection{Combined Representations}
\seclabel{comb-rep}


Approaches with \emph{combined} representations aim to leverage the advantages of multiple representation types and mitigate their limitations. Most combine volumetric and surface representations to utilise information on both voxel observation states and the scene geometry.

\citet{Kriegel2012, Kriegel2015} present a combined approach with a probabilistic voxel grid by extending earlier work using a surface representation \citep{Kriegel2011}. Views are proposed to extend the boundaries of a surface mesh and are assigned an \gls{inf} value from the voxel grid.

\citet{Song2018} and \citet{Song2020a} extend earlier work on a volumetric approach \citep{Song2017} with a surface representation. The surface representation is obtained by creating a Poisson reconstruction \citep{Kazhdan2006} from a \gls{tsdf} extracted from the voxel grid. A trajectory is planned between views sampled with RRT* to observe uncertain surfaces in the reconstruction and exploration frontier voxels. \cite{Song2021} present a similar approach for online \gls{mvs} reconstruction.    

\citet{Low2007} represent scene observations with a combination of voxels and surface patches. Their approach aims to obtain a minimum measurement density within each occupied voxel. Measurements in voxels with insufficient density are connected to define surface patches. Views are proposed to observe these patches and a next best view is selected to observe the patch with the greatest potential increase in measurement density.

\citet{Karaszewski2016} present a multistage combined approach. The first stage, originally presented by \citet{Karaszewski2012}, uses a volumetric representation with a density-based measurement classification to propose and select views of scene regions with insufficient measurements. A Poisson surface reconstruction is created from this initial observation and then refined by capturing more views.

\citet{Dierenbach2016} use the \gls{gng} algorithm \citep{Fritzke1995} to learn a model of the scene geometry from sensor measurements. This model defines a graph of connected nodes which partition the scene volume into a Voronoi tessellation, where each node lies at the centre of a Voronoi cell. Views are proposed to observe each cell and a next best view is selected to observe the cell with the lowest measurement density.

\citet{Monica2018a} present an approach with a surfel representation. Surfels that lie on the boundary between unobserved and unoccupied voxels are classified as frontels. Next best views are selected to observe the set of visible frontels with the greatest surface area.

\citet{Ran2023} present an approach that uses a novel \gls{nerf}-based representation and estimates the neural uncertainty of implicit surface points. Views are sampled by RRT* and next best views are chosen to reduce the neural uncertainty.  

Approaches with combined representations can obtain more complete and accurate scene observations than other approaches by leveraging the advantages of multiple representations; however, a greater computational cost is typically incurred for maintaining several representations. 

\subsection{Pointcloud Representations}
\seclabel{free-rep}

Many of the limitations associated with structured scene representations can be overcome by using an \emph{unstructured} pointcloud representation. These directly represent the observation state using sensor measurements instead of encoding scene information in an external structure. They do not reduce the fidelity of scene information and avoid some of the computational costs (e.g., raycasting) incurred by maintaining an external structure.  

\gls{see} is the first \gls{nbv} approach to use a fully unstructured representation, to the best of our knowledge, but other approaches have since been presented. \citet{Peralta2020,Zeng2020a} train neural networks to select views from a predefined set that can obtain the greatest improvements in surface coverage. \citet{Arce2020} present a multistage approach with a density-based measurement classification similar to \gls{see}. \citet{Williams2020a} use the \gls{hpr} algorithm \citep{Katz1969} to identify point-based frontiers on the boundaries of visible surfaces.

\gls{see} is a measurement-direct \gls{nbv} approach with a density-based pointcloud representation. All scene information is directly associated with sensor measurements and therefore \gls{see} avoids the computational cost of maintaining an external structure (i.e., a voxel grid or surface mesh). The computational cost of updating the density representation scales with the number of measurements captured and processed at the discrete views chosen by \gls{see} rather than the scene volume or the resolution of an external structure.

Measurements are processed when they are added to the observation and are only reprocessed if they have insufficient density and lie within the classification neighbourhood of newly added measurements. Visibility and occlusion checking is only required for the small subset of points used to propose views for extending an observation, instead of for every added measurement as is typically required for volumetric approaches.   

The fidelity of scene information is not constrained by a structural resolution since captured measurements are individually classified based on the density of neighbouring points instead of being aggregated into a single set of values for each unit of the external structure (e.g., a voxel). \gls{see} leverages this detailed knowledge to identify scene regions that require further observation, propose views that avoid occlusions and select next best views that can obtain the best improvements in surface coverage. This enables it to observe scenes with less observation time and often higher completion than structured approaches.     

\section{The Surface Edge Explorer (SEE)}
\seclabel{see}

This paper presents \gls{see}, a \gls{nbv} planning approach with a density-based pointcloud representation. \gls{see} aims to obtain complete scene observations by capturing a minimum measurement density from all visible surfaces. Sensor measurements are individually classified by the number of neighbouring points within a resolution radius \psecref{set-params}. Measurements with a minimum number of neighbours are \emph{core} points and those without are \emph{outlier} points. Outliers with core neighbours become \emph{frontier} points, which in the context of this work represent a boundary in the captured measurements between fully and partially observed surfaces.

This density-based classification of measurements is used to identify the boundaries between sufficiently and insufficiently observed surfaces \psecref{class-points}. Views are then proposed to observe the frontier points that define these boundaries \psecref{prop-views}. Known occlusions are handled proactively by detecting occluding points before a view is obtained and proposing an alternative unoccluded view \psecref{opt-views}. The visibility of surfaces from views is quantified by encoding the shared visibility of frontier points from views in a graphical representation \psecref{view-vis}. Next best views are chosen from this graph to obtain significant improvements in surface coverage while reducing travel distance \psecref{sel-nbv}. If a target frontier point is not observed from a view --- typically due to an unknown occlusion or surface discontinuity --- then it is reactively adjusted to avoid the obstruction \psecref{adj-views}. Views are captured until there are no frontiers remaining \psecref{obs-comp}.   

\begin{algorithm}[tpb]
	\caption{SEE($\viewcurr, \Gamma_\mathrm{see}, \Gamma_\mathrm{sensor}$)}
	\begin{algorithmic}[1]
		\small
		\State {$\Pall \equiv \core \cup \frontier \cup \free$}	
		\State {$\Pnew, \core, \frontier, \free, \Vprop, \mathcal{G} = \emptyset$}
		\State {$\frontp_\mathrm{c}\; \gets \algnull;\; \Vobs \gets \algnull;\; \surfgeom \gets \algnull $}	
		\State {$\mathrm{ComputeParameters}(\Gamma_\mathrm{see}, \Gamma_\mathrm{sensor})$} 
		\Repeat
		\State {$\mathrm{GetNewMeasurements}(\viewcurr, \Pnew)$}
		\State {$\mathrm{ClassifyPoints}(\viewcurr, \Pnew, \core, \frontier, \free, \Vobs)$}
		\If {$\frontp_\mathrm{c} \in \frontier$}
		\State {$\mathrm{AdjustView}(\viewcurr, \frontp_\mathrm{c}, \Pnew, \frontier, \free, \Vprop, \Vobs, \surfgeom)$}
		\EndIf
		\State {$\mathrm{ProposeViews}(\viewcurr, \Pall, \Pnew, \frontier, \Vprop, \surfgeom)$}
		\State {$\mathrm{RefineViews}(\viewf_\mathrm{c}, \Pall, \frontier, \free, \Vprop, \Vobs, \surfgeom)$}
		\State {$\mathrm{GraphViews}(\viewcurr, \frontier, \Vprop, \mathcal{G})$}
		\State {$\mathrm{SelectNBV}(\viewf_\mathrm{c}, \frontp_\mathrm{c}, \mathcal{G})$}
		\Until {$\frontier = \emptyset$}
		\State \Return {$\mathrm{complete}$}
	\end{algorithmic}
	\algmlabel{see-alg}
\end{algorithm}

\talgmref{see-alg} presents an overview of \gls{see}. The observation parameters, $\Gamma_\mathrm{see}$, are set using user-specified values and the sensor properties, $\Gamma_\mathrm{sensor}$ (Line 4). New measurements are iteratively obtained and processed to select a next best view (Line 5). A set of new measurements, $\Pnew$, is captured from a sensor at the current view, $\viewcurr$ (Line 6). These measurements are added to the \gls{see} pointcloud, $\Pall$, and the classifications of core, $\core$, frontier, $\frontier$, and outlier, $\free$, points are updated. The current  view position is also recorded in the captured view set, $\Vobs$, for future occlusion handling (Line 7). If the target frontier point, $\frontp_\mathrm{c}$, associated with the current view was not successfully observed (i.e., it is still classified as a frontier point) then the view is adjusted (Lines 8--10). New views are proposed to observe the new frontier points and added to the set of view proposals, $\Vprop$. The local surface geometry around each new frontier is then estimated, $\surfgeom$ (Line 11). The nearest view proposals to the current view are proactively checked for occlusions and refined if necessary (Line 12). The connectivity of these views in the frontier visibility graph, $\mathcal{G}$, is then updated by evaluating their shared visibility of frontier points (Line 13). A next best view is selected from this graph and the sensor is moved to capture new measurements (Line 14). The observation completes when there are no frontiers remaining (Lines 15--16).

\subsection{Computing Suitable Parameters}
\seclabel{set-params}

\begin{table}
	\caption{The configuration parameters of \gls{see}.}
	\centering
	\begin{tabular}{@{}lll@{}}
		\toprule
		Parameter & Description & Units\\ \midrule
		$\rho$ & Target measurement density & points per m$^3$\\
		$r$                  & Resolution radius & m\\
		$d$                   & View distance & m\\
		$\epsilon$            & Minimum separation distance & m\\
		$\psi$                & Occlusion search distance & m\\
		$\upsilon$            & Visibility search distance & m\\
		$\tau$                & Maximum views to update  & number of views\\
		\bottomrule
	\end{tabular}
	\tbllabel{see-param-table}
\end{table} 

\begin{algorithm}[tpb]
	\caption{ComputeParameters($\Gamma_\mathrm{see}, \Gamma_\mathrm{sensor}$)}
	\mathleft
	\begin{algorithmic}[1]
		\small
		\State {$\Gamma_\mathrm{see} \equiv (r, \rho, d, \epsilon, k_\mathrm{min})$}
		\State {$\Gamma_\mathrm{sensor} \equiv (\omega_\mathrm{x}, \omega_\mathrm{y}, \theta_\mathrm{x}, \theta_\mathrm{y})$}
		\If {$r = 0 \algand \rho \neq 0$}
		\State {$\displaystyle r \gets \left(\frac{9}{4\pi\rho}\right)^\frac{1}{3}$}
		\EndIf
		\If {$\rho = 0 \algand d \neq 0 \algand r \neq 0$}
		\State {$\displaystyle\rho \gets {\frac{\omega_\mathrm{x}\omega_\mathrm{y}}{4\tan{0.5\theta_\mathrm{x}}\tan{0.5\theta_\mathrm{y}}(3d^2+2r^2)}}$}
		\EndIf
		\If {$d = 0 \algand \rho \neq 0 \algand r \neq 0$}
		\State {$\displaystyle d \gets \sqrt{\frac{\omega_\mathrm{x}\omega_\mathrm{y}}{12\rho\tan{0.5\theta_\mathrm{x}}\tan{0.5\theta_\mathrm{y}}} - \frac{2r^2}{3}}$}
		\EndIf
		\If {$\epsilon = 0$}
		\State {$\displaystyle\epsilon \gets \left(\frac{3r}{2\pi\rho}\right)^\frac{1}{3}$}
		\EndIf
		\State {$\displaystyle k_\mathrm{min} \gets \left\lceil\frac{4}{3}\pi\rho r^3\right\rceil$}
	\end{algorithmic}
	\mathcenter
	\algmlabel{set-params}
\end{algorithm}

\gls{see} aims to obtain scene observations with a minimum measurement density over all visible surfaces. This measurement density, $\rho$, is calculated within an $r$-radius sphere. Sensor measurements are captured at a view distance, $d$, and separated by a minimum distance, $\epsilon$. These parameters \ptblref{see-param-table} can be user-specified or computed from other parameters and the sensor properties. It is also useful to have knowledge of the measurement noise and scene scale in order to set suitable values for an observation, but detailed scene-specific information is not required. 

The target measurement density, $\rho$, determines how many measurements need to be captured from scene surfaces. It should be set sufficiently high to attain the desired level of structural detail from a scene. The resolution radius, $r$, defines the scale at which the measurement density is evaluated. This should be large enough to handle measurement noise robustly (i.e., most noisy measurements that deviate from true surfaces should be encompassed within the resolution radius to prevent them from being classified as frontier points) while still retaining surface features and computational efficiency. 

The view distance, $d$, sets the range at which a sensor observes the scene. It should be set so that sufficient frontier points can be identified from captured measurements. If the view distance is too large, the measurements will be sparsely distributed over scene surfaces and may all be outlier points. When the sensor is too close, the measurements will be densely distributed over surfaces and may all be core points. A method is presented to compute a suitable distance from the measurement density and sensor properties (Alg. 2, Line 10). The minimum separation distance, $\epsilon$, between measurements is used to reduce memory consumption and computational cost. It should be set small enough that it does not affect point classification. 

\talgmref{set-params} presents the calculation of these parameters from the sensor resolution, $\omega_\mathrm{x}$ and $\omega_\mathrm{y}$, and field-of-view, $\theta_\mathrm{x}$ and $\theta_\mathrm{y}$. If the user specifies a value for the target density but not the resolution radius then it is computed such that the resulting volume will contain three points at the target density (Lines 3--5). If the resolution radius and view distance parameters are set, but the target density is not, then it is calculated from the sensor properties to equal the density of sensor measurements that could be captured from the largest observable surface area at the specified view distance (Lines 6--8). If the target density and resolution radius are set but not the view distance then it is computed such that the measurement density captured from the largest observable surface would be equal to the target density (Lines 9--11).

If the minimum separation distance is not specified then a suitable value is computed from the target density and resolution radius (Lines 12--14). The final parameter, $k_\mathrm{min}$, is not user-configurable and defines the number of points that need to exist in an $r$-radius sphere with density $\rho$ (Line 15). 

The remaining user-specified parameters tune the practical performance of \gls{see}. The occlusion search distance, $\psi$, is the radius around a frontier point that is searched for occluding points. It should ideally be equal to the view distance to capture all occluding points but can be reduced to limit computational cost. The visibility search distance, $\upsilon$, is the radius around the sight line of a view that is searched for occluding points. It should be large enough that an unoccluded view is able to successfully observe new measurements around its associated frontier. The maximum views to update, $\tau$, defines the number of neighbouring views that are processed when handling occlusions and updating the frontier visibility graph. It should be as high as possible without incurring a significant computational cost.

\subsection{Processing Sensor Measurements}
\seclabel{class-points}

Sensor measurements are classified based on the number of neighbouring points within the resolution radius \figref{class}. Measurements with more neighbours than the target density are classified as \emph{core} points and those without are \emph{outlier} points. Outlier points with core neighbours are \emph{frontier} points. These frontiers define a boundary between sufficiently and insufficiently observed surfaces. 


\begin{figure}[tpb]
	\centering
	\includegraphics[width=\linewidth]{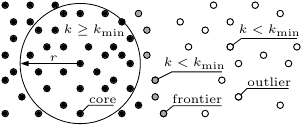}
	\caption{An illustration of the density-based classification used by \gls{see} \palgmref{class-points}. Measurements with a sufficient number of neighbours, $k_\mathrm{min}$, in an $r$-radius are classified as core (black) while those without are outliers (white). Outlier measurements with core neighbours are frontiers (grey).}
	\figlabel{class}
\end{figure}

New sensor measurements are added to the \gls{see} pointcloud and assigned a classification. This density-based classification is based on \acrlong{dbscan} \citep[\acrshort{dbscan};][]{Ester1996}\glsunset{dbscan}. Sensor measurements, $\Pall := \{\mathbf{p}_i\}_{i=1}^n$, where $\mathbf{p}_i \in \mathbb{R}^3$, are classified as core points, $\core$, frontier points, $\frontier$, or outlier points, $\free$. The point classifications are complete and unique, so every point is assigned to a single class; i.e., 
\begin{equation*}
P = \core \cup \frontier \cup \free
\end{equation*} 
\vspace{-1ex}
and
\vspace{-1ex}
\begin{equation*}
\core \cap \frontier = \core \cap \free = \frontier \cap \free = \emptyset\,.
\end{equation*}
The set of measurements, $N_\mathbf{p}$, in the pointcloud within an $r$-radius of a point, $\mathbf{p}$, is given by 
\begin{equation*}
N_\mathbf{p} := \mathrm{Neighbours}_r(P,r,\mathbf{p}) := \{\mathbf{q} \in P \;|\;||\mathbf{q} - \mathbf{p}|| \leq r\}\,,
\end{equation*} 
where $||\cdot||$ denotes the L$^2$-norm.

A point is classified as core if it has at least $k_\mathrm{min}$ neighbours, 
\begin{equation*}
\core := \{\mathbf{p} \in P \;|\; |N_\mathbf{p}| \geq k_\mathrm{min}\}\,,
\end{equation*}
where $|\cdot|$ denotes set cardinality.

A point is classified as a frontier if it has fewer than $k_\mathrm{min}$ neighbours, some of which are core points, 
\begin{equation*}
	\frontier := \{\mathbf{p} \in P \;|\; |N_\mathbf{p}| < k_\mathrm{min} \,\land\, N_\mathbf{p} \,\cap\, \core \not= \emptyset \}\,,
\end{equation*}
or as an outlier otherwise,
\begin{equation*}
\free := \{\mathbf{p} \in P \;|\; |N_\mathbf{p}| < k_\mathrm{min} \,\land\, N_\mathbf{p} \,\cap\, \core = \emptyset\}\,.
\end{equation*}

\begin{algorithm}[tpb]
	\caption{ClassifyPoints($\viewcurr, \Pnew, \core, \frontier, \free, \Vobs$)}
	\mathleft
	\begin{algorithmic}[1]
		\small
		\State {$\viewidx{\mathrm{c}};\; \Pall \equiv \core \cup \frontier \cup \free$}
		\State {$\visited \gets \emptyset$}
		\ForAll{{$\mathbf{p} \in \Pnew$}}
		\If{{$\mathrm{Neighbours}_r(\Pall,\epsilon,\mathbf{p}) = \emptyset$}}
		\State {$\Pall \gets \Pall \cup \{\mathbf{p}\}$}
		\State $Q \gets \{\mathbf{p}\} \cup \mathrm{Neighbours}_r(P,r,\mathbf{p})$
		\State{$\Vobs[\mathbf{p}] \gets \viewpcurr$}
		\ForAll{$\mathbf{q} \in Q$}
		\If{{$\mathbf{q} \notin \core$}}
		\State $N_\mathbf{q} \gets \mathrm{Neighbours}_r(P,r,\mathbf{q})$
		\If{$|N_\mathbf{q}| < k_\mathrm{min}$}
		\If{{$N_\mathbf{q} \cap \core \neq \emptyset$}}
		\State $\frontier \gets \frontier \cup \{\mathbf{q}\}$
		\If { $\mathbf{q} \in \free$}
		\State { $\free \gets \free \setminus \{\mathbf{q}\}$}
		\EndIf
		\Else
		\State $\free \gets \free \cup \{\mathbf{q}\}$
		\EndIf
		\Else
		\State $\core\gets \core\cup \{\mathbf{q}\}$
		\If { $\mathbf{q} \in \frontier$}
		\State { $\frontier \gets \frontier \setminus \{\mathbf{q}\}$}
		\EndIf
		\If { $\mathbf{q} \in \free$}
		\State { $\free \gets \free \setminus \{\mathbf{q}\}$}
		\EndIf
		\If{$\mathbf{p} \not= \mathbf{q} \algand \mathbf{q} \notin \visited$}
		\State {$Q \gets Q \cup N_\mathbf{q}$}
		\State {$\visited \gets \visited \cup \{\mathbf{q}\}$}
		\EndIf
		\EndIf
		\EndIf
		\EndFor
		\EndIf
		\EndFor
	\end{algorithmic}
	\algmlabel{class-points}
	\mathcenter
\end{algorithm}

\talgmref{class-points} presents the classification of new measurements. Each point in the new measurement set, $\mathbf{p} \in \Pnew$, is processed (Line 3). Any point that satisfies the $\epsilon$-radius constraint is added to the pointcloud, $\Pall$, and the (re)classification queue, $Q$, along with its neighbourhood points (Lines 4--6). The current view position, $\viewpcurr$, is stored in $\Vobs$ for future occlusion handling (Line 7). 

If a point in the queue is not a core point then it is (re)classified based on the new measurements (Lines 8--10). Points with insufficient neighbours to be core are (re)classified as frontier points if they have core neighbours or become outliers (Lines 11--19). Points with sufficient neighbours are (re)classified as core points (Lines 20--27). If an unprocessed point is (re)classified as a core point then its neighbours are added to the (re)classification queue and it is marked as processed (Lines 28--31). The classification procedure completes when all new measurements and those in the (re)classification queue are processed.

Classifying sensor measurements based on the density of neighbouring points distinguishes a boundary between scene regions that are completely observed (i.e., consist only of core points) from those that require additional measurements (i.e., contain frontier and outlier points). The frontier points identified along this boundary are used to propose views for extending the coverage of completely observed surfaces.

\subsection{Proposing Views}
\seclabel{prop-views}

Observation coverage is improved by capturing measurements from surfaces around frontier points. Views are proposed to observe the frontiers by estimating the local surface geometry from an eigendecomposition of measurements within their $r$-radius neighbourhoods. The surface geometry is described by a set of orthogonal vectors \psecref{surf-geom}. They represent the local surface normal, a boundary between complete and incomplete surfaces and the direction of partial observation (i.e., a frontier vector). The outwards facing normal direction is determined by evaluating the visibility of vectors pointing in both potential directions \psecref{norm-dir}.

Each view, $\view$, is proposed to observe the local surface around a frontier point, $\frontp$ \figref{vp}. The view position, $\viewp$, is set at the view distance, $d$, along the surface normal and the view orientation, $\viewo$, is given by the surface normal, $\norm$.

\begin{figure}[tpb]
	\centering
	\includegraphics[width=0.9\linewidth]{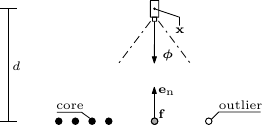}
	\caption{An illustration of a view proposal \palgmref{prop-views}. A view, $\view$, is proposed to observe the surfaces around a target frontier point (grey dot), $\frontp$, using the locally estimated surface normal, $\norm$. The view position, $\viewp$, is set at the view distance, $d$, from the frontier point along the normal vector, $\norm$. The view orientation, $\viewo$, is the negative direction of the normal vector.}
	\figlabel{vp}
\end{figure}

\begin{algorithm}[tpb]
	\caption{ProposeViews($\viewcurr, \Pall, \Pnew, \frontier, \Vprop, \surfgeom$)}
	\mathleft
	\begin{algorithmic}[1]
		\small
		\State{$\viewidx{\mathrm{c}};\;\surfgeom[\frontp] \equiv (\norm, \front, \bound)$}
		\ForAll {$\frontp \in \frontier \algand \Vprop[\frontp] = \algnull$}
		\State{$\mathrm{EstimateSurface}(\viewcurr, \frontp, \Pall, \Pnew, \surfgeom)$}
		\State{$\Vprop[\frontp] \gets (\frontp + d\norm, -\norm)$}
		\EndFor
	\end{algorithmic}
	\mathcenter
	\algmlabel{prop-views}
\end{algorithm}



\talgmref{prop-views} presents the generation of a view proposal for each new frontier point, $\frontp$ (Line 2). Measurements within an $r$-radius of the frontier point, including the frontier, are processed to estimate the local surface geometry, $\surfgeom[\frontp]$ (Line 3; Alg. \ref{alg:est-geom}). A view is generated and associated with the target frontier in the view proposal set, $\Vprop[\frontp]$ (Line 4).

\subsubsection{Estimating the Surface Geometry}
\seclabel{surf-geom}

\begin{figure}[tpb]
	\centering
	\includegraphics[width=\linewidth]{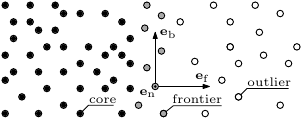}
	\caption{An illustration of the orthogonal vectors that represent the local surface geometry \palgmref{est-geom}. The normal vector, $\norm$, is normal to the local plane (out of page), the frontier vector, $\front$, points towards the region of partial observation, and the boundary vector, $\bound$, lies along the boundary between the fully and partially observed regions.}
	\figlabel{surf}
\end{figure}

The surface geometry around a frontier point is estimated by an eigendecomposition of the neighbouring measurements within an $r$-radius. This produces a planar estimate defined by three orthogonal vectors, each of which describes one component of the local surface geometry \figref{surf}.  

\talgmref{est-geom} presents the estimation of local surface geometry. A planar estimate of the local geometry around a frontier point, $\mathbf{f}$, is computed from a matrix representation, $\mathbf{D} \in \mathbb{R}^{3 \times |N_\mathbf{f}|}$, of its neighbours, $N_\frontp$ (Lines 1--2). A covariance matrix, $\mathbf{A}$, is computed from the neighbourhood matrix and an eigendecomposition is performed to produce a set of eigenvalues, $\Lambda$, and associated eigenvectors, $\Upsilon$, that satisfy the eigenequation (Lines 3--5).


Each eigenvector describes one component of the estimated surface geometry. The normal vector, $\norm$, is orthogonal to the surface plane. The frontier vector, $\front$, lies on the surface plane and points in the direction of partial observation. The boundary vector, $\bound$, points along the border between partially and fully observed surfaces. The eigenvectors are assigned based on their eigenvalues.

The normal vector, $\norm$, points along the axis with the least variation in neighbouring measurements. It is the eigenvector with the minimum eigenvalue (Line 6). The correct direction for the normal vector (i.e., outwards from the surface) is determined by the visibility of vectors pointing in both potential directions from the current view (Line 7; Alg. \ref{alg:norm-dir}).   

The frontier vector, $\front$, points towards the partially observed region of the scene. It is the remaining eigenvector whose dot product with the mean of the neighbouring measurements, $\mathbf{{\bar{p}}}$, is greatest (Lines 8--9). The vector direction is defined to have a positive dot product with a vector from the mean point to the frontier so it is oriented towards the partially observed region of the scene (Line 10).

The boundary vector, $\bound$, is locally tangential to the border between the density regions and is defined by the cross product of the normal and frontier vectors (Line 11). The three orthogonal vectors are then stored in $\surfgeom$ (Line 12).

\begin{algorithm}[tpb]
	\caption{EstimateSurface($\viewcurr, \frontp, \Pall, \Pnew, \surfgeom$)}
	\mathleft
	\begin{algorithmic}[1]
		\small
		\State {$N_\frontp \gets \mathrm{Neighbours}_r(P,r,\frontp) \cup \{\frontp\}$}
		\State{$\mathbf{D} \gets [\mathbf{p}_{1}-\mathbf{f},...,\mathbf{p}_{n}-\mathbf{f}] \algwhere \mathbf{p}_{i} \in N_\mathbf{f}$}
		\State{$\mathbf{A} \gets \mathbf{DD}^\mathrm{T}$}
		\State {$\Lambda \gets \mathrm{eigenvalues}(\mathbf{A}) \algwhere \Lambda \equiv (\lambda_{1},\lambda_{2},\lambda_{3})$}
		\State {$\Upsilon \gets \mathrm{eigenvectors}(\mathbf{A}) \algwhere \Upsilon \equiv (\boldsymbol{\upsilon}_{1}, \boldsymbol{\upsilon}_{2}, \boldsymbol{\upsilon}_{3})$}
		\State{$\norm \gets \{\boldsymbol{\upsilon}_i \;|\; \lambda_i = \min\left\lbrace\Lambda\right\rbrace\}$}
		\State{$\mathrm{DirectNormal}(\viewcurr, \frontp, \norm, \Pnew)$}
		\State{$\displaystyle\mathbf{{\bar{p}}} \gets \frac{1}{|N_\mathbf{f}|}\sum_{\mathbf{p} \in N_\mathbf{f}}{\left(\frontp - \mathbf{p}\right)}$}
		\State{$\displaystyle\front \gets \argmax_{\boldsymbol{\upsilon}_i \,\in\, \Upsilon \setminus \norm} (|\mathbf{\bar{p}} \cdot \boldsymbol{\upsilon}_i|)$}
		\State{$\front \gets \mathrm{sign}(\mathbf{\bar{p}} \cdot \front)\front$}
		\State{$\bound \gets \norm \times \front$}
		\State{$\surfgeom[\frontp] \gets (\norm, \front, \bound)$}
	\end{algorithmic}
	\mathcenter
	\algmlabel{est-geom}
\end{algorithm} 


\subsubsection{Determining the Correct Normal Direction}
\seclabel{norm-dir}

\begin{figure}[tpb]
	\centering
	\includegraphics[width=0.95\linewidth]{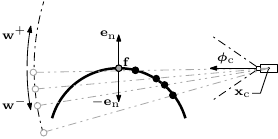}
	\caption{An illustration of how the correct normal direction is determined \palgmref{norm-dir}. Normal vectors pointing in opposite directions, $\norm$ and $-\norm$, have their visibility from the current view, $\viewidx{\mathrm{c}}$, evaluated. The correct normal vector will not be occluded by surface measurements closer to the current view (black dots). This vector is found by projecting the occluding measurements onto a sphere and searching along both projected vectors, $\mathbf{w}^+$ and $\mathbf{w}^-$, until free space is found. }
	\figlabel{norm-dir}
\end{figure}

The views proposed to observe frontier points are defined by their associated surface normals. These normals must point outwards from the surface in order to obtain valid views. 

The direction of a normal can often be defined as pointing towards the current view. This technique works well when the current view is close to the surface normal but can fail when it observes the surface at an acute angle. In this scenario, the sightline and surface normal are nearly perpendicular and measurement noise can corrupt the identification of the outwards facing normal direction.

A more robust method to determine the outwards facing normal direction is to evaluate the visibility of both potential vectors from the current view \figref{norm-dir}. The normal direction pointing outwards from the surface will not be occluded by measurements while the other direction will be.

The visibility of the two potential normal vectors from the current view is evaluated by sampling points along each vector, starting at the frontier point, and checking for an unoccluded sight line. The visibility of points is evaluated in both directions until the first unoccluded point is found. Its corresponding vector defines the outwards facing normal.

The visibility of points is evaluated by projecting them onto the surface of a unit sphere centred on the current view, a method inspired by \gls{hpr} \citep{Katz1969}. Measurements captured from the current view that are closer to the view than the normal vectors are also projected. This projection preserves the relative orientation of points to the view while normalising the distance. Projected points with similar sight lines are close to each other on the sphere surface and a point is considered occluded if projected sensor measurements exist within a specified radius.

\begin{algorithm}[tpb]
	\caption{DirectNormal($\viewcurr, \frontp, \norm, \Pnew$)}
	\mathleft
	\begin{algorithmic}[1]
		\small
		\State{$\viewidx{\mathrm{c}}$}
		\State{$\mathbf{w}^+ \gets \frontp - \viewpcurr$}
		\State{$\mathbf{w}^- \gets \mathbf{w}^+$}
		\Repeat 
		\State{$\mathbf{w}^+ \gets \mathbf{w}^+ + \upsilon\norm$}
		\State{$\mathbf{w}^- \gets \mathbf{w}^- - \upsilon\norm$}
		\State{$\displaystyle J \gets \left\{\frac{\mathbf{p} - \viewpcurr}{||\mathbf{p} - \viewpcurr||} \;\bigg|\; \mathbf{p} \in \Pnew \land ||\mathbf{p} - \viewpcurr|| < ||\mathbf{w}^+||\right\}$}
		\State{$\displaystyle N_{\mathbf{w}^+} \gets \mathrm{Neighbours}_r\left(J, \upsilon, \frac{\mathbf{w}^+}{||\mathbf{w}^+||}\right)$}
		\State{$\displaystyle N_{\mathbf{w}^-} \gets \mathrm{Neighbours}_r\left(J, \upsilon, \frac{\mathbf{w}^-}{||\mathbf{w}^-||}\right)$}
		\Until {$N_{\mathbf{w}^+} = \emptyset \algor N_{\mathbf{w}^-} = \emptyset$}
		\If {$N_{\mathbf{w}^+} \neq \emptyset \algand N_{\mathbf{w}^-} = \emptyset$}
		\State {$\norm \gets -\norm$}
		\EndIf
	\end{algorithmic}
	\mathcenter
	\algmlabel{norm-dir}
\end{algorithm}

\talgmref{norm-dir} presents the calculation of the outwards facing normal direction. Sampled points along each vector, $\mathbf{w}^+$ and $\mathbf{w}^-$, are initialised to the position of the frontier point relative to the current view position, $\frontp - \viewpcurr$ (Lines 2--3). These sampled points are iteratively moved along the normal vectors until a search for projected measurements on the sphere in either direction returns an empty set (Line 4). 

In each iteration, the sampled points are moved along their respective vectors by the visibility search distance, $\upsilon$ (Lines 5--6). New measurements that are closer to the current view than the sampled points are projected onto the surface of a unit sphere centred on the current view position, $\viewpcurr$, to create a projected set, $J$ (Line 7). The sampled points are then projected onto the sphere surface and the set of projected measurements is searched for any occluding points within an $\upsilon$-radius of the projected samples (Lines 8--9). The outwards facing normal direction is found when one of the search result sets, $N_{\mathbf{w}^+}$ or $N_{\mathbf{w}^-}$, is empty (Line 10). The normal is reversed if the negative normal direction set is empty, $N_{\mathbf{w}^-} = \emptyset$, and the positive set is not (Lines 11--13).

This method is able to reliably determine the outwards facing normal direction for the estimated surface around a frontier point. It improves observation efficiency by identifying the correct direction from which to propose a view and reduces the number of failed views attempted.     

\subsection{Refining Views}
\seclabel{opt-views}

Proposed views can only observe their target frontier points if they are unoccluded. Occlusions are handled \emph{proactively} by identifying occluded views before they are visited \psecref{occ-detect}. These occluded views are refined to alternative unoccluded views \psecref{occ-propose}.

\talgmref{opt-views} presents the proactive identification of known occlusions and the resulting view refinement. The $\tau$-nearest views, $N_{\viewcurr}$, to the current sensor position, $\viewpcurr$, are selected from the view proposal set, $\Vprop$, and processed (Lines 3--5). If occluding measurements are found (Line 6; Alg. \ref{alg:occ-detect}) then an optimisation strategy is used to identify an unoccluded view of the frontier (Line 7; Alg. \ref{alg:opt-view}). If an unoccluded view is not found the frontier is reclassified as an outlier (Lines 8--10); otherwise, the existing view is replaced (Lines 11--13).            

\begin{algorithm}[tpb]
	\mathleft
	\caption{RefineViews($\viewf_\mathrm{c}, \Pall, \frontier, \free, \Vprop, \Vobs, \surfgeom$)}
	\begin{algorithmic}[1]
		\small
		\State {$\viewidx{\mathrm{c}}$}
		\State {$\viewopt \gets \algnull$}
		\State {$N_{\viewcurr} \gets \mathrm{Neighbours}_k(\Vprop, \tau, \viewpcurr)$}
		\ForAll {$\viewf \in N_{\viewcurr}$}
		\State {$\frontp \gets \frontp \in \frontier \;\;\mathrm{s.t.}\;\; \viewf = \Vprop[\frontp]$}
		\If {$\mathrm{IsOccluded}(\viewf, \frontp, \Pall, \surfgeom)$}
		\State {$\mathrm{OptimiseView}(\viewopt, \frontp, \Pall, \Vobs, \surfgeom)$}
		\If {$\mathrm{IsOccluded}(\viewopt, \frontp, \Pall, \surfgeom)$}
		\State{$\frontier \gets \frontier \setminus \frontp$}
		\State{$\free \gets \free \cup \frontp$}
		\Else
		\State {$\Vprop[\frontp] \gets \viewopt$}
		\EndIf
		\EndIf
		\EndFor
	\end{algorithmic}
	\mathcenter
	\algmlabel{opt-views}
\end{algorithm}

\begin{figure}[tpb]
	\centering
	\includegraphics[width=0.9\linewidth]{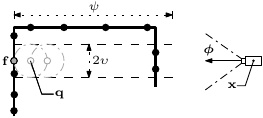}
	\caption{An illustration of detecting occlusions between a frontier point and its associated view \palgmref{occ-detect}. Measurements (black dots) within a set distance of the sight line are assumed to represent occluding surfaces. They are found by performing an $\upsilon$-radius search around discrete points (grey circles), $\qpoint \in Q$, up to a distance, $\psi$, from the frontier (grey dot).}
	\figlabel{occ-search}
\end{figure}

\subsubsection{Detecting Occlusions}
\seclabel{occ-detect}

A frontier point is occluded if measurements exist within the visibility search distance, $\nu$, of the sight line from the target view. Occluding measurements are found by searching the $\upsilon$-radius neighbourhoods of points sampled along the sight line \figref{occ-search}. A frontier is visible from a view if no occluding measurements are found.

\talgmref{occ-detect} presents the method for occlusion detection between a frontier point and a view. Points are sampled along the sight line at an $\upsilon$-interval, starting at an offset from the frontier, $\zeta$, and ending at the occlusion search distance, $\psi$ (Lines 3--5). The $\upsilon$-radius neighbourhood around every sampled point is searched for occluding measurements (Line 6). The view is occluded if the union of the search result sets, $N_\qpoint$, is not empty (Line 7).

The search for occluding measurements starts at an offset from the frontier to account for measurement noise. A suitable offset is identified by searching along the local surface normal until the first region of free space is found. 

\talgmref{vis-off} presents the calculation of this offset. Points are incrementally sampled along the normal vector at an $\upsilon$-interval, starting at the frontier point, until the $\upsilon$-radius neighbourhood around the newest sampled point is empty or the occlusion search distance is reached (Lines 2--6).

Proactively detecting known occlusions improves observation efficiency by reducing the number of unsuccessful views. This method also aids in the selection of next best views that can observe more frontiers by quantifying the shared visibility of frontiers between views.   

\begin{algorithm}[tpb]
	\mathleft
	\caption{IsOccluded($\viewf, \frontp, \Pall, \surfgeom$)}
	\begin{algorithmic}[1]
		\small
		\State {$\view$}
		\State {$\zeta \gets 0$}
		\State {$\mathrm{GetVisibilityOffset}(\frontp, \zeta, \Pall, \surfgeom) $}
		\State {$\displaystyle \viewo_\mathrm{s} = \frac{\frontp - \viewp}{||\frontp - \viewp||}$}
		\State {$\displaystyle Q \gets \left\{\frontp - i\viewo_\mathrm{s} \;|\; i = \zeta, \zeta + \upsilon, \dots, \psi \right\}$}
		\State {$\displaystyle N_\qpoint \gets \bigcup\limits_{\qpoint \in Q} \mathrm{Neighbours}_r(\Pall, \upsilon, \qpoint)$}
		\State \Return {$N_\qpoint \neq \emptyset$}
	\end{algorithmic}
	\mathcenter
	\algmlabel{occ-detect}
\end{algorithm}

\begin{algorithm}[tpb]
	\mathleft
	\caption{GetVisibilityOffset($\frontp, \zeta, \Pall, \surfgeom$)}
	\begin{algorithmic}[1]
		\small
		\State{$\surfgeom[\frontp] \equiv (\norm, \front, \bound)$}
		\Repeat
		\State {$\zeta \gets \zeta + \upsilon$}
		\State {$\qpoint \gets \frontp + \zeta\norm$}
		\State {$N_\qpoint \gets \mathrm{Neighbours}_r(\Pall, \upsilon, \qpoint)$}
		\Until {$N_\qpoint = \emptyset \algor \zeta \geq \psi$}
	\end{algorithmic}
	\mathcenter
	\algmlabel{vis-off}
\end{algorithm}

\subsubsection{Proposing an Unoccluded View}
\seclabel{occ-propose}

Occluded views are updated to an unoccluded view by finding the sight line to their frontier point that has the greatest separation from any occluding measurements. This sight line provides the best chance that the frontier will be successfully observed.  

A frontier point is occluded when measurements exist along the sight line between it and the proposed view. A frontier's visibility from different views can be found by projecting possibly occluding measurements onto a unit sphere \figref{occ-repocc}, using the same method discussed in \tsecref{norm-dir}. The sphere centre may be slightly offset from the frontier to account for measurement noise. The projection preserves the direction of occluding measurements while placing them on a uniform manifold for efficient processing.

An unoccluded view of the frontier point is found by solving a \emph{maximin} optimisation problem on the spherical projection. This maximises the minimum distance between the processed view and any of the projected points. The maximin problem solution is the antipole \citep{Wesolowsky1983} of the complementary \emph{minimax} problem solution \citep{Patel2002}.  

\begin{figure}[tpb]
	\centering
	\includegraphics[width=0.6\linewidth]{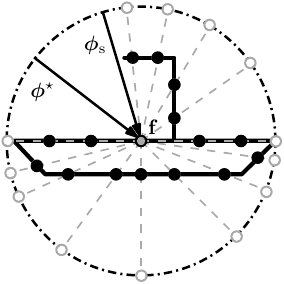}
	\caption{A 2D illustration of the spherical projection used to find an unoccluded sight line of a target frontier point (grey dot), $\frontp$ \palgmref{opt-view}. Measurements (black dots) are projected onto a unit sphere (grey circles) centred on the frontier point. The optimally unoccluded view orientation, $\viewoopt$, is maximally separated from the projected points. The sight line from the capturing view, $\viewoobs$, is known to be unoccluded.}
	\figlabel{occ-repocc}
\end{figure} 

The minimax solution is the centre of the smallest spherical cap that contains the projected points. This cap is defined by a plane intersecting the sphere and is found by optimising the plane normal and its distance from the sphere centre. The minimax solution is the intersection point of the plane normal with the sphere.

The specific optimisation method depends on the distribution of projected points. If they are spread over more than a hemisphere then the smallest containing cap is found by minimising the distance of the plane from the sphere centre. If the projected points lie on less than a hemisphere then the smallest containing cap is found by maximising the distance of the plane from the sphere centre.

\begin{algorithm}[tpb]
	\mathleft
	\caption{OptimiseView($\viewopt, \frontp, \Pall, \Vobs, \surfgeom$)}
	\begin{algorithmic}[1]
		\small
		\State {$\zeta \gets 0$}
		\State {$\mathrm{GetVisibilityOffset}(\frontp, \zeta, \Pall, \surfgeom) $}
		\State {$\displaystyle\viewoobs \gets \frac{\frontp - \Vobs[\frontp]}{||\frontp - \Vobs[\frontp]||}$}
		\State {$\cpoint \gets \frontp - \zeta\viewoobs$}
		\State {$\displaystyle J \gets \left\lbrace \frac{\point-\cpoint}{||\point-\cpoint||} \,\middle|\, \point \in \mathrm{Neighbours}_r(P,\psi, \frontp) \right\rbrace$\vspace{-2ex}}
		\State {\parbox{\linewidth}{\begin{argmini*}
					{\substack{\npoint\,\in\,\mathbb{R}^3,\,e\,\in\,[0,1]}}{e}
					{}{(\npoint^\star,e^\star)\gets}
					\addConstraint{e}{\leq \npoint^T\npoint\,}
					\addConstraint{e}{\geq \npoint^T\mathbf{j}\,,}{\;\;\forall\;\mathbf{j} \in J}
					\addConstraint{\hspace{-13ex}\mathrm{initial\;condition}\quad}{\!\!\!\npoint(0)= -\viewo_\mathrm{s}}
			\end{argmini*}}\vspace{-2ex}}
		\If{$e^\star \not= 0$}
		\State{$\displaystyle\viewoopt \gets -\frac{\npoint^\star}{||\npoint^\star||}$}
		\Else
		\State{\hspace{-4ex}\parbox{\linewidth}{\begin{argmaxi*}
					{\substack{\npoint\,\in\,\mathbb{R}^3,\,e\,\in\,[0,1]}}{e}
					{}{\quad\;\;(\npoint^\star,e^\star)\gets}
					\addConstraint{e}{\geq \npoint^T\npoint\,}
					\addConstraint{e}{\leq \npoint^T\mathbf{j}\,,}{\;\;\forall\;\mathbf{j} \in J}
					\addConstraint{\hspace{-13ex}\mathrm{initial\;condition}\quad}{\!\!\!\npoint(0)= \viewo_\mathrm{s}}
			\end{argmaxi*}}\vspace{-2ex}}
		\State{$\displaystyle\viewoopt \gets \frac{\npoint^\star}{||\npoint^\star||}$}
		\EndIf
		\State {$\viewpopt \gets \frontp - d\viewoopt$}
		\State {$\viewopt \gets (\viewpopt, \viewoopt)$}
	\end{algorithmic}
	\mathcenter
	\algmlabel{opt-view}
\end{algorithm}

\talgmref{opt-view} presents the calculation of an unoccluded view using this maximin optimisation strategy. The projection centre of the sphere, $\cpoint$, is offset from the frontier point, $\frontp$, by the visibility offset \palgmref{vis-off}, $\zeta$, towards the capturing view position, $\Vobs[\frontp]$, as this sight line is known to be unoccluded (Lines 2--4). Neighbouring measurements within the occlusion search distance, $\psi$, of the frontier are projected onto a unit sphere (Line 5). The projected points, $J$, are initially assumed to be distributed over more than a hemisphere and the full sphere optimisation is performed (Line 6). The distance of a plane intersecting the sphere from its centre is minimised while ensuring the plane normal satisfies the optimisation constraints and the projected points all lie on the same side of the plane. The plane normal is initialised to the opposite direction of the sight line from the capturing view, $\viewoobs$, as this is known to be unoccluded. The optimised view orientation, $\viewoopt$, points in the opposite direction of the optimised normal so the sphere is intersected at the maximin solution (Line 8).

If the projected points lie on less than a hemisphere then the full sphere optimisation converges to a plane bisecting the sphere (i.e., $e^\star = 0$) and a hemispherical optimisation is performed (Lines 9--10). The distance of a plane intersecting the sphere from its centre is maximised while ensuring the plane normal satisfies the optimisation constraints and the projected points all lie on the same side of the plane. The plane normal is initialised to the direction of the capturing view orientation as this is known to be unoccluded. The optimised view orientation points in the same direction as the optimised normal (Line 11). After the view optimisation is complete the optimised view position, $\viewpopt$, is set at the view distance, $d$, from the frontier in the opposite direction of the optimised view orientation and the optimised view proposal, $\viewopt$, is set (Lines 13--14). 

This optimisation is guaranteed to find a view of a frontier point that is free from known occlusions if one exists. Refining proposed views with it improves the efficiency of scene observations by increasing the chance that a frontier point will be successfully observed from its associated view.

\subsection{Quantifying Views}
\seclabel{view-vis}

\begin{algorithm}[tpb]
	\mathleft
	\caption{GraphViews($\viewcurr, \frontier, \Vprop, \mathcal{G}$)}
	\begin{algorithmic}[1]
		\small
		\State{$\mathcal{G} \equiv (M, K);\; \fvpair;\; \viewidx{\mathrm{c}};\; \viewidx{i}$}
		\State{$M^\prime \gets \{(\frontp, \Vprop[\frontp]) \;|\; \frontp \in \frontier \}$}
		\State{$K^\prime \gets K \setminus \{(\fvpairm_a, \,\fvpairm_b) \;|\; \frontp_a \notin \frontier \lor \frontp_b \notin \frontier\}$}
		\State {$N_{\viewcurr} \gets \mathrm{Neighbours}_k(\Vprop, \tau, \viewpcurr)$}
		\ForAll {$\viewf_i \in N_{\viewcurr}$}
		\State {$\frontp_i \gets \frontp_i \in \frontier \;\;\mathrm{s.t.}\;\; \viewf_i = \Vprop[\frontp_i]$}
		\State{$K^\prime \gets K^\prime \setminus \{(\fvpairm_a, \,\fvpairm_b) \;|\; \frontp_a = \frontp_i\}$}
		\State {$N_{\viewf_i} \gets \mathrm{Neighbours}_k(\Vprop, \tau, \viewp_i)$}
		\ForAll {$\viewf_j \in N_{\viewf_i}$}
		\State {$\frontp_j \gets \frontp_j \in \frontier \;\;\mathrm{s.t.}\;\; \viewf_j = \Vprop[\frontp_j]$}
		\If {$\neg\mathrm{IsOccluded}(\frontp_j, \viewf_i, \psi, \upsilon)$}
		\State {$K^\prime \gets K^\prime \cup \{(\fvpairm_i, \,\fvpairm_j)\}$}
		\EndIf
		\EndFor
		\EndFor
		\State {$\mathcal{G} = (M^\prime, K^\prime)$}
	\end{algorithmic}
	\mathcenter
	\algmlabel{view-vis}
\end{algorithm}  

\begin{figure}[tpb]
	\centering
	\includegraphics[width=0.85\linewidth]{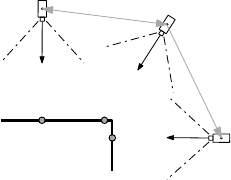}
	\caption{An illustration of the frontier visibility graph \palgmref{view-vis}. The connectivity between frontier-view pairs (grey dots and sensors) is represented with directed edges (grey arrows). An edge from a parent to a child denotes that the child frontier point is visible from the parent view proposal.}
	\figlabel{vis-fgph}
\end{figure}

The best views are those that capture the most new surface coverage. \gls{see} quantifies the predicted coverage from a proposed view as the number of visible frontier points. The visibility of frontiers from views is encoded in a directed \emph{frontier visibility graph} that connects each frontier to the views from which it can be observed. It is used to select next best views that can obtain large increases in surface coverage. 

The frontier visibility graph, $\mathcal{G} := (M, K)$, represents the view proposals and their visible frontier points \figref{vis-fgph}. The graph vertices, $M$, are pairs, $\fvpair$, of a frontier point, $\frontp$, and its associated view, $\viewf$. An edge, $(\fvpairm_i, \,\fvpairm_j) \in K$, exists from a parent vertex, $\fvpairm_i$, to a child vertex, $\fvpairm_j$, if the parent view, $\viewf_i$, can observe the child frontier point, $\frontp_j$. This covisibility is calculated using occlusion detection \psecref{occ-detect} and quantifies the expected number of frontiers visible from each view.

\talgmref{view-vis} presents the calculation of the frontier visibility graph. The vertices, $M'$, are a new set of frontier-view pairs created from the view proposal set, $\Vprop$ (Line 2). A new edge set, $K'$, is created from the existing edge set, $K$, by removing edges between vertices that have been reclassified as core points (Line 3). Only the set of frontier-view pairs corresponding with the $\tau$-nearest view proposals, $N_{\viewcurr}$, to the current sensor position, $\viewpcurr$, are processed (Line 4). 

Each frontier-view pair, $\fvpairm_i = \fvpairnmidx{i}$, is processed by removing existing outgoing edges from the graph and adding new outgoing edges to the visible frontier points associated with the $\tau$-nearest view proposals, $N_{\viewf_i}$ (Lines 5--10). The visibility of these frontier points is evaluated with occlusion detection (Line 11; Alg. \ref{alg:occ-detect}). An outgoing edge is added from the processed vertex, $\fvpairm_i$, to the associated vertex, $\fvpairm_j$, if its frontier is visible (Line 12). The new frontier visibility graph is the new vertex and edge sets when all of the queued frontier-view pairs have been processed (Line 16).           

The frontier visibility graph quantifies views that should obtain large increases in surface coverage. This information is used by the next best view selection metric to choose views that can capture an efficient scene observation.     

\subsection{Selecting a Next Best View}
\seclabel{sel-nbv}

\begin{algorithm}[tpb]
	\mathleft
	\caption{SelectNBV($\viewf_\mathrm{c}, \frontp_\mathrm{c}, \mathcal{G}$)}
	\begin{algorithmic}[1]
		\small
		\State{$\mathcal{G} \equiv (M, K);\; \fvpair;\; \view;\; \viewidx{\mathrm{c}}$}
		\State{$\displaystyle\fvpairm^\prime \gets \argmin_{\fvpairm \in M}\left(||\viewp - \viewp_\mathrm{c}||\right)$}
		\State{$\displaystyle M^\prime \gets \{\fvpairm \in M \,|\, (\fvpairm,\,\fvpairm^\prime) \in K \land \mathrm{deg}^+(\fvpairm) > \mathrm{deg}^+(\fvpairm^\prime) \}$}
		\State {$\displaystyle\fvpairm^\star \gets \argmax_{\fvpairm \in M^\prime}\left(\frac{\mathrm{deg}^+(\fvpairm)}{||\viewp - \viewp_\mathrm{c}||}\right)$}
		\If {$\fvpairm^\star = \textsc{null}$}
		\State {$\displaystyle\fvpairm^\star \gets \fvpairm^\prime$}
		\EndIf
		\State {$(\frontp_\mathrm{c}, \viewf_\mathrm{c}) \gets \fvpairm^\star$}
	\end{algorithmic}
	\mathcenter
	\algmlabel{sel-nbv}
\end{algorithm}

\begin{figure}[tpb]
	\centering
	\includegraphics[width=0.75\linewidth]{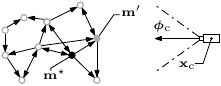}
	\caption{An illustration of the next best view selection metric \palgmref{sel-nbv}. Vertices (grey circles) in the frontier visibility graph represent frontier-view pairs and are connected with directed edges denoting visibility (black arrows). The sensor represents the current view, $\viewidx{\mathrm{c}}$. The next best view is the vertex (black dot), $\fvpairm^\star$, that has the greatest outdegree relative to its view's distance from the sensor position, $\viewp_\mathrm{c}$, with an edge to the vertex (grey dot), $\fvpairm^\prime$, whose view is closest to the sensor.}
	\figlabel{vis-lrat}
\end{figure}

Next best views are selected to improve a scene observation while reducing the acquisition cost. \gls{see} selects views with the most visible frontier points relative to their distance from the current sensor position. This ratio penalizes views far from the current sensor position that can not observe more frontiers than closer views. Euclidean distance may differ significantly from the actual travel distance of the sensor on some platforms, but \gls{see} is platform agnostic and it therefore provides the best available estimate of travel cost between sensor positions.     

This greedy view selection behaviour may select a distant view with many visible frontier points that then requires the sensor to return to capture closer surfaces. This problem is avoided by requiring the chosen view to have visibility of the frontier point associated with the closest view proposal \figref{vis-lrat}. The closest view can have a very small travel distance so it is only selected when it has visibility of at least as many frontiers as any other potential view.

\talgmref{sel-nbv} presents the selection of next best views. The closest view proposal, $\fvpairm^\prime$, to the current sensor position, $\viewp_\mathrm{c}$, is used to define the set of permissible views (Lines 2--3). The vertex with the greatest ratio between the number of outgoing edges (i.e., its outdegree) and the distance of its view proposal from the current sensor position, $\fvpairm^\star$, is chosen as the next view (Line 4). If no such view exists the closest view is selected (Lines 5--7). The next best view and frontier point are then updated to the chosen view (Line 8).     

This next best view selection metric chooses views that can capture significant improvements in scene coverage while travelling short distances. It enables \gls{see} to obtain observations with a low travel distances by capturing local surfaces before travelling to larger unobserved regions.  

\subsection{Handling Failed Views}
\seclabel{adj-views}

Views do not always successfully observe their associated frontiers. A view is considered failed if its associated frontier is not reclassified as a core point after processing the newly captured measurements. This occurs when part of the local surface is not visible, either due to occlusions or surface discontinuities (e.g., corners). These failed views must be adjusted to successfully observe their target frontiers.

A new view is proposed by using the captured measurements to compute an adjustment for the failed view. This adjustment aims to avoid occlusions and surface discontinuities by reducing the separation distance between the frontier and the mean of the captured measurements. It is calculated from translations along, and rotations around, the local surface geometry vectors \figref{view-adj}.

The view adjustment is repeated until the target frontier is successfully observed or a termination criterion is reached. When the separation between the frontier point and the mean of the captured measurements stops decreasing the adjusted view is reinitialized to the position from which the frontier point was first captured, as this is known to be unoccluded. This new view is again adjusted until the frontier is successfully observed or the process is terminated. If this process also fails then the frontier is reclassified as an outlier.

\begin{figure}[tpb]
	\centering
	\includegraphics[width=0.8\linewidth]{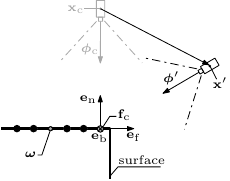}
	\caption{An illustration of a view adjustment around a surface discontinuity \palgmref{adj-view}. The frontier point, $\frontp_\mathrm{c}$, on one side of a surface edge can not be successfully observed by the current view, $\viewidx{\mathrm{c}}$. An adjusted view, $\viewprime{\prime}$, with visibility around this discontinuity is calculated from the difference between the frontier and the mean (grey dot) of the measurements (black dots) captured from the current view, $\boldsymbol{\omega}$.}
	\figlabel{view-adj}
\end{figure}

\talgmref{adj-view} presents the view adjustment procedure. The adjustment parameters are initialised if this is the first time the view has been adjusted (Lines 3--7). Individual adjustments are calculated using a scaling factor, $A[\frontp_\mathrm{c}]$, that increases with each subsequent adjustment. An adjustment is applied if the current separation distance between the frontier and the mean of the captured measurements is less than the previous separation distance, $D[\frontp_\mathrm{c}]$. The view switch flag, $V_\mathrm{switch}[\frontp_\mathrm{c}]$, indicates when the adjustment of a view has failed and it has been reset to the frontier's capturing view.

\begin{algorithm}[tpb]
	\caption{AdjustView($\viewcurr, \frontp_\mathrm{c}, \Pnew, \frontier, \free, \Vprop, \Vobs, \surfgeom$)}
	\mathleft
	\begin{algorithmic}[1]
		\small
		\State{$\viewidx{\mathrm{c}};\; \viewprime{\prime}$}
		\State{$\surfgeom[\frontp_\mathrm{c}] \equiv (\norm, \front, \bound);\; \frontd \equiv [s_0, s_1, s_2]^\mathrm{T}$}
		\If{$D[\frontp_\mathrm{c}] = \algnull$}
		\State{$D[\frontp_\mathrm{c}] \gets \infty$}
		\State{$A[\frontp_\mathrm{c}] \gets 1$}
		\State {$V_\mathrm{switch}[\frontp_\mathrm{c}] \gets \textsc{false}$}
		\EndIf
		\State{$\mathbf{C} \gets [\norm\;\front\;\bound]$}
		\State{$\displaystyle\cmass \gets \dfrac{1}{|\Pnew|}\sum_{\mathbf{p} \in \Pnew}{\mathbf{p}}$}
		\State{$\frontd \gets \mathbf{C}^\mathrm{T}(\frontp_\mathrm{c} - \cmass)$}
		\If {$||\frontd|| < D[\frontp_\mathrm{c}]$}
		\State{$\mathbf{t}_\mathrm{f} \gets (A[\frontp_\mathrm{c}] + 1)\frontdi_1\front$}
		\State{$\mathbf{t}_\mathrm{b} \gets (A[\frontp_\mathrm{c}] + 1)\frontdi_2\bound$}
		\State {$\displaystyle\theta_\mathrm{b} = \tan^{-1}\left(\frac{dA[\frontp_\mathrm{c}]\frontdi_1}{d^2 + (A[\frontp_\mathrm{c}] + 1)\frontdi_1^2}\right)$}
		\State {$\displaystyle\theta_\mathrm{f} = \tan^{-1}\left(\frac{dA[\frontp_\mathrm{c}]\frontdi_2}{d^2 + (A[\frontp_\mathrm{c}] + 1)\frontdi_2^2}\right)$}
		\State{$\mathbf{R}_\mathrm{b} \gets \mathbf{I} + (\sin\theta_\mathrm{b})\bound^\times  + (1 - \cos\theta_\mathrm{b})(\bound^\times)^2$}
		\State{$\mathbf{R}_\mathrm{f} \gets \mathbf{I} + (\sin\theta_\mathrm{f})\front^\times  + (1 - \cos\theta_\mathrm{f})(\front^\times)^2$}
		\State{$\viewo^\prime \gets \frontp_\mathrm{c} - \mathbf{R}_\mathrm{f}(\mathbf{t}_\mathrm{b} + \mathbf{R}_\mathrm{b}(\mathbf{t}_\mathrm{f} + \viewpcurr))$}
		\State{$D[\frontp_\mathrm{c}] \gets ||\frontd||$}
		\State{$A[\frontp_\mathrm{c}] \gets 2A[\frontp_\mathrm{c}]$}
		\ElsIf {$\neg V_\mathrm{switch}[\frontp_\mathrm{c}]$}
		\State {$\viewo^\prime \gets \frontp_\mathrm{c} - \Vobs[\frontp_\mathrm{c}]$}
		\State{$D[\frontp_\mathrm{c}] \gets \infty$}
		\State{$A[\frontp_\mathrm{c}] \gets 1$}
		\State {$V_\mathrm{switch}[\frontp_\mathrm{c}] \gets \textsc{true}$}
		\Else
		\State{$\frontier \gets \frontier \setminus \frontp_\mathrm{c}$}
		\State{$\free \gets \free \cup \frontp_\mathrm{c}$}
		\EndIf
		\If {$\frontp_\mathrm{c} \in \frontier$}
		\State{$\displaystyle\viewp^\prime \gets \frontp_\mathrm{c} - d\frac{\viewo^\prime}{||\viewo^\prime||}$}
		\State{$\Vprop[\frontp_\mathrm{c}] \gets \viewf^\prime$}
		\EndIf 
	\end{algorithmic}
	\mathcenter
	\algmlabel{adj-view}
\end{algorithm}

The view adjustment is performed in a coordinate frame, $\mathbf{C}$, defined by the local surface geometry (Lines 8--9). The current separation distance, $\frontd = [s_0, s_1, s_2]^\mathrm{T}$, between the frontier, $\frontp_\mathrm{c}$, and the mean of the captured measurements, $\cmass$, is calculated and the view is only adjusted if it is less than the previous separation distance (Lines 10--11).

The translational adjustments, $\mathbf{t}_\mathrm{f}$ and $\mathbf{t}_\mathrm{b}$, along each axis, $\front$ and $\bound$, are the scaled separation distances (Lines 12--13). They move the centre of the viewing frustum towards regions where the previous view captured no measurements. This moves the view around discontinuities at the intersection of different surfaces (i.e., corners). 

The rotational adjustments, $\mathbf{R}_\mathrm{b}$ and $\mathbf{R}_\mathrm{f}$, for each axis are computed by Rodrigues' rotation formula \citep{RodriguesO.1840a}, \[\mathbf{R} = \mathbf{I} + (\sin\theta)\mathbf{u}^\times  + (1 - \cos\theta)(\mathbf{u}^\times)^2\,,\] where $\mathbf{I}$ is the identity matrix, $\theta$ is the angle of rotation, $\mathbf{u}$ is the axis of rotation, and $(\cdot)^\times$ is the skew symmetric matrix of a vector, \[
\begin{bmatrix}
	u_0 \\ u_1 \\ u_2
\end{bmatrix}^{\!\!\times} = \begin{bmatrix}
	0       & -u_{2} & u_{1} \\
	u_{2}       & 0 & -u_{0} \\
	-u_{1}       & u_{0} & 0
\end{bmatrix}\,.\] These rotations move the view around newly discovered occlusions and towards the unobserved side of surface discontinuities (Lines 14--17).

The adjusted view orientation, $\viewo^\prime$, is the vector pointing towards the frontier point from the newly translated and rotated view position (Line 18). The separation distance parameter is updated in case further adjustments are required (Line 19). The adjustment scaling factor is doubled to prevent the magnitude of future adjustments from converging to zero as the separation distance decreases (Line 20).  

A view adjustment is terminated when the separation distance stops decreasing. If the adjustment started from the initial view proposed then the new view orientation, $\viewo^\prime$, is the sight line from the frontier's capturing view, as this is known to be occlusion-free (Lines 21--22). The adjustment parameters are reinitialised as this \emph{switched} view is also adjusted reactively if it is unsuccessful (Lines 23--25). If adjustment of the switched view also fails then the frontier is reclassified as an outlier (Lines 26--29).

The new position, $\viewp^\prime$, of an adjusted or switched view is set at the view distance from the frontier along the new view direction (Line 30--31). The new view, $\viewf^\prime$, then replaces the existing one in the set of proposed views (Line 32).

This reactive adjustment of failed views enables scene coverage to be extended beyond surface discontinuities and previously unseen occluding surfaces. It enables \gls{see} to obtain highly complete observations by capturing measurements from surfaces with restricted visibility.   

\subsection{Completing an Observation}
\seclabel{obs-comp}

A scene observation completes when there are no remaining frontiers and all measurements are classified as either core or outlier points. The extent of a scene observation can be bounded by discarding all points outside of a given volume. 

\section{Simulation Experiments}
\seclabel{exp}

The simulation experiments compare the observation performance of \gls{see} with several volumetric \gls{nbv} planning approaches on numerous models of varying size and geometric complexity. All of the evaluated approaches were tuned to capture the highest surface coverage using the least views and have a worst case computation time of less than ten seconds per view. The volumetric approaches and other versions of \gls{see} have been evaluated on some of the models in previous work \citep{Delmerico2017,Border2018,BorderThesis,Border2019} but these results are not directly comparable due to differences in the model sizes, simulated platforms, simulated sensors and parameters (e.g., the view distance). These changes were necessary as the experiments in this paper were designed to simulate realistic platforms with limited observation time budgets. 

\gls{see} is compared with seven volumetric \gls{nbv} planning approaches: Average Entropy \citep[AE;][]{Kriegel2015}, Area Factor \citep[AF;][]{Vasquez-Gomez2014}, Occlusion Aware \citep[OA;][]{Delmerico2017}, Proximity Count \citep[PC;][]{Delmerico2017}, Rear Side Entropy \citep[RSE;][]{Delmerico2017}, Rear Side Voxel \citep[RSV;][]{Delmerico2017} and Unobserved Voxel \citep[UV;][]{Delmerico2017}. The implementations of these volumetric approaches are provided by \citet{Delmerico2017}. 

Experiments were performed with six small-scale models: Newell Teapot \citep{Newell1975}, Stanford Bunny \citep{Turk1994}, Stanford Dragon \citep{Curless1996}, Stanford Armadillo \citep{Krishnamurthy1996a}, Happy Buddha \citep{Curless1996} and Helix \citep{helix}, and three large-scale models: Statue of Liberty \citep{liberty}, Radcliffe Camera \citep{radcliffe} and Notre-Dame de Paris \citep{notredame}. The algorithms were run for 100 independent experiments on each model. 

The small models were observed in a robot arm simulation environment. This consisted of a UR10 robot arm with an RGB-D camera attached to the end effector and a turntable \figref{ur10-env}. The turntable centre and UR10 base are separated by $0.75$~m. The turntable has a diameter of $0.8$~m, so the small models are scaled to fit within a $0.8$x$0.8$x$0.6$~m bounding box. The maximum model height is $0.6$~m so that views above a model are reachable by the end effector.   

The large models were observed in an aerial simulation environment. Measurements are captured by a LiDAR \citep[i.e., a similar configuration to Hovermap;][]{Hudson2022} mounted onto the underside of a quadrotor with a two-axis gimbal. The large models are placed at the origin on a virtual ground plane and scaled to fit within a $40$x$40$x$40$~m box. The quadrotor is able to reach any collision-free view position and the two-axis gimbal can position the LiDAR at any view orientation in the hemisphere below the quadrotor. 

The simulated sensors are defined by a field-of-view in degrees, $\theta_\mathrm{x}$ and $\theta_\mathrm{y}$, and a resolution in pixels, $w_\mathrm{x}$ and $w_\mathrm{y}$ \ptblref{sensor-table}. Sensor measurements were obtained by raycasting into the triangulated surface of a model and adding Gaussian noise ($\mu = 0$~m, $\sigma = 0.01$~m) to the ray intersections. This noise magnitude was chosen to be representative of the noise associated with measurements from a depth camera (e.g., an Intel RealSense L515) and a LiDAR (e.g., a Hesai XT32). 

Collision-free paths between views are planned with \acrlong{aitstar} \citep[\acrshort{aitstar};][]{Strub2020,Strub2022}\glsunset{aitstar}, using the \acrlong{ompl} \citep[\acrshort{ompl};][]{ompl}\glsunset{ompl}, and executed with MoveIt \citep{Coleman2014}. The platform-specific reachability of each next best view is evaluated before planning a path and any unreachable view is adjusted to a reachable view based on platform-specific constraints. If the path planning to a view fails then the \gls{nbv} algorithm is forced to select a different view.       

\gls{see} selects next best views until its completion criterion is satisfied. The volumetric approaches use a view limit termination criterion that is set by the user. For each model, the volumetric algorithms are limited to the largest number of views taken by \gls{see} for any run on the model. This provides a fair comparison with \gls{see} by ensuring they have sufficient views to achieve highly complete observations of the models.

View proposals for the volumetric approaches are sampled from a surface encompassing the scene, in this case a hemisphere, as presented by \citet{Vasquez-Gomez2014} and \citet{Delmerico2017}. \citet{Kriegel2015} does not sample views from an encompassing view surface but we use the implementation provided by \citet{Delmerico2017} which does. The radius of the view hemisphere is set to the sum of the view distance, $d$, and a surface offset equal to the mean distance of points in the model from the origin. The number of views sampled from the hemisphere is $2.4$ times the view limit, as presented by \citet{Delmerico2017}.

\begin{figure}[tpb]
	\centering
	\includegraphics[width=0.75\linewidth]{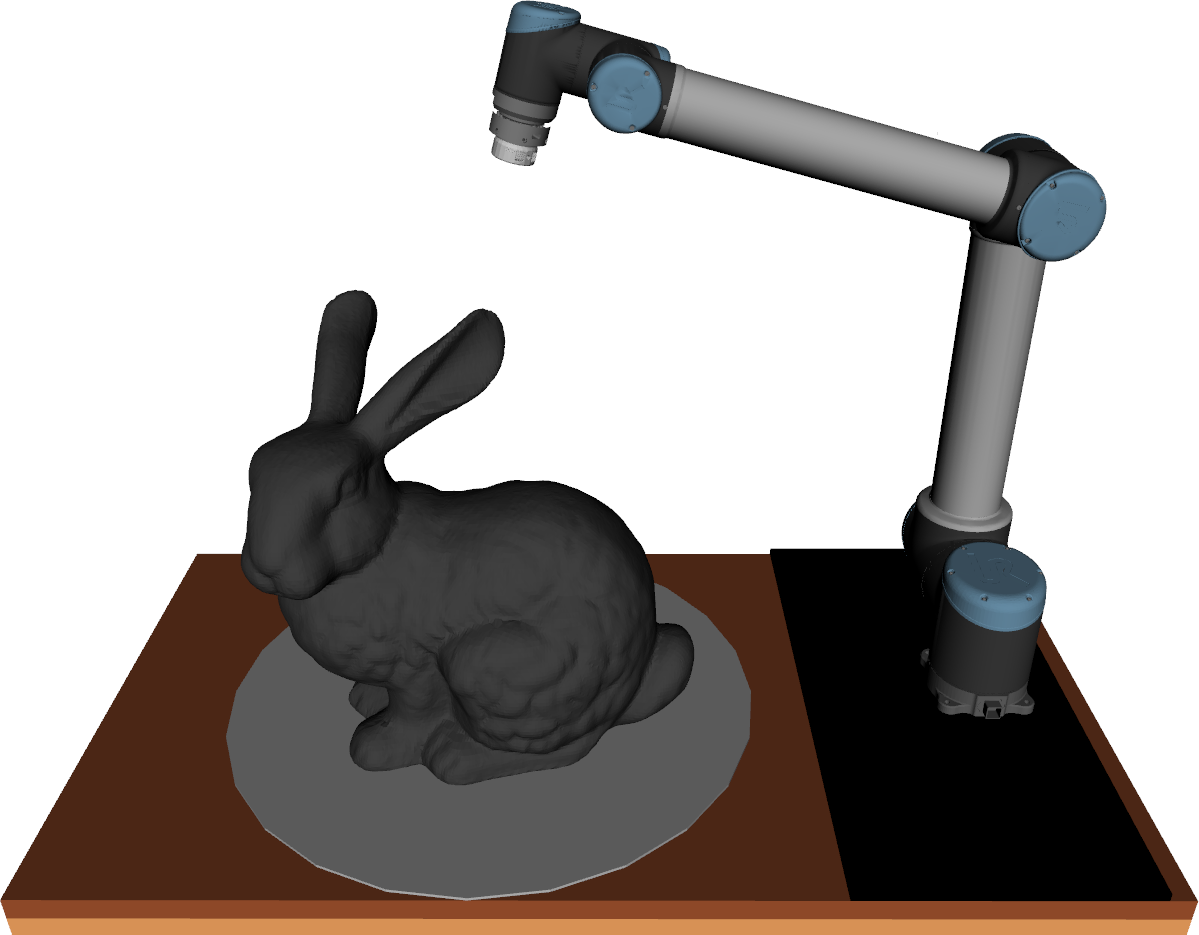}
	\caption{The UR10 simulation environment used in the experiments in \tsecref{exp}. An object (dark grey) is placed on the turntable (light grey) and measurements are captured with an RGB-D camera attached to the UR10 end effector.}
	\figlabel{ur10-env}
\end{figure}

\begin{table}
	\caption{The field-of-view in degrees, $\theta_\mathrm{x}$ and $\theta_\mathrm{y}$, and resolution in pixels, $w_\mathrm{x}$ and $w_\mathrm{y}$, of the RGB-D camera and LiDAR used to capture sensor measurements in the experiments in \tsecref{exp}.}
	\centering
	\begin{tabular}{@{}llll@{}}
		\toprule
		Property & RGB-D Camera & LiDAR & Units \\ \midrule
		$\theta_\mathrm{x}$ & $70$        & $60$ & degrees\\
		$\theta_\mathrm{y}$ & $43$        & $40$ & degrees\\
		$w_\mathrm{x}$      & $848$       & $1200$ & pixels\\
		$w_\mathrm{y}$      & $480$       & $800$ & pixels\\ \bottomrule
	\end{tabular}
	\tbllabel{sensor-table}
\end{table}  

\subsection{Performance Metrics}
\seclabel{see-metrics}

The observation performance of each approach is quantified by the surface coverage obtained, travel distance required and time used when capturing an observation. These values are averaged across the $100$ independent experiments performed on each model with each approach.

The surface coverage is calculated from the mesh vertices of the 3D model, $\Mall$. A model vertex is considered covered when the captured pointcloud has a measurement within a user-selected registration radius, $\eta$ \ptblref{param-table}. The surface coverage obtained by an algorithm is then measured as the ratio of covered model points, \[\Mobs = \{\mathbf{s} \in \Mall \;|\;\exists \mathbf{p} \in \Pall, \;||\, \mathbf{s} - \mathbf{p} \,|| \leq \eta\}\,,\] to total model points,
\[\mathrm{Coverage} = \frac{|\Mobs|}{|\Mall|}\,.\] 

The sensor travel distance is measured as the summed lengths of the paths travelled by the sensor between views. The total observation time is the time required to process new sensor measurements, select a next best view, plan a collision-free path to the chosen view and move the sensor from its current position to the next best view.

\subsection{Algorithm Parameters}

\ttblref{param-table} presents the parameters used by \gls{see} and the evaluated volumetric approaches to observe the small- and large-scale scene models. The minimum separation distance between sensor measurements used by \gls{see}, $\epsilon$, is also applied to the volumetric approaches to reduce memory consumption and computational cost for all the evaluated approaches and ensure a fair comparison. 

\subsubsection{Small Models}
\seclabel{small-models}

The target measurement density, $\rho$, for \gls{see} on the small models is computed from the resolution radius, view distance and sensor properties \palgmlineref{set-params}{Line 4}. The resolution radius, $r$, is set large enough to robustly handle measurement noise and is also used as the voxel size for the volumetric approaches. The view distance, $d$, is chosen to be far enough that a significant proportion of the scene is visible from each view while remaining reachable by the UR10. The occlusion search distance, $\psi$, is set equal to the view distance so that all known occlusions can be identified. The visibility search distance, $\upsilon$, is set slightly smaller than the resolution radius so that frontier points on narrow concave surfaces (e.g., between the folded segments of the Stanford Dragon) can be accurately evaluated. The view update limit, $\tau$, is chosen to process a large number of proposed views while maintaining a reasonable computational cost. The values of these parameters are presented in \ttblref{param-table}.

The raycasting resolution parameter, $\xi$, sets the fraction of the sensor resolution that is raycast by volumetric approaches to calculate the \gls{inf} value of view proposals. It is chosen to be small enough to attain a worst case computation time of less than ten seconds per view without unduly impacting the observation performance. The travel cost weight, $\gamma$, for the volumetric approaches is set to zero so that they obtain the highest surface coverage possible within the specified view limit. 

\subsubsection{Large Models}
\seclabel{large-models}

The target density for \gls{see} on the large models is set to capture small surface details. The resolution radius is chosen to be large enough to accurately estimate the local surface geometry and is again also used as the voxel size for the volumetric approaches. The view distance is computed from these parameters and the sensor properties \palgmlineref{set-params}{Line 6}. The occlusion search distance is set to half the model size to reduce the computational cost. The visibility search distance is set to the resolution radius as this is sufficiently small to determine the visibility of frontier points on concave surfaces in the large models (e.g., the balconies on the Radcliffe Camera). The view update limit is set as large as possible while maintaining a reasonable computational cost. The values of these parameters are presented in \ttblref{param-table}. 

The raycasting resolution for the volumetric approaches is chosen to attain a worst case computation time of less than ten seconds per view and their travel cost weight is set to zero for the highest surface coverage.

\begin{table}
	\caption{The parameters used for \gls{see}, the volumetric approaches and the associated analysis in the experiments in \tsecref{exp}. \underline{Underlined} values were derived from other parameters.}
	\centering
	\begin{tabular}{@{}llll@{}}
		\toprule
		& Small Models & Large Models & Units \\ \midrule
		$\rho$ & \underline{490738} & $300$ & points per m$^3$ \\
		$r$ & $0.03$ & $0.15$ & m \\
		$d$ & $0.5$ & \underline{35.6} & m \\
		$\epsilon$ & \underline{0.003} & \underline{0.06} & m \\
		$\psi$ & $0.5$ & $20$ & m \\
		$\upsilon$ & $0.01$ & $0.15$ & m \\
		$\tau$ & $100$ & $100$ & number of views \\
		$\xi$ & $0.1$ & $0.01$ & \\
		$\gamma$ & $0$ & $0$ & \\
		$\eta$ & $0.005$ & $0.05$ & m \\
		\bottomrule
	\end{tabular}
	\tbllabel{param-table}
\end{table}  

\begin{figure*}[tpb]
	\centering
	\captionsetup[subfigure]{}
	\captionsetup[subfigure]{labelformat=empty}
	\captionsetup[subfigure]{justification=centering}
	
	\subfloat[Newell Teapot]{\includegraphics[width=.32\linewidth]{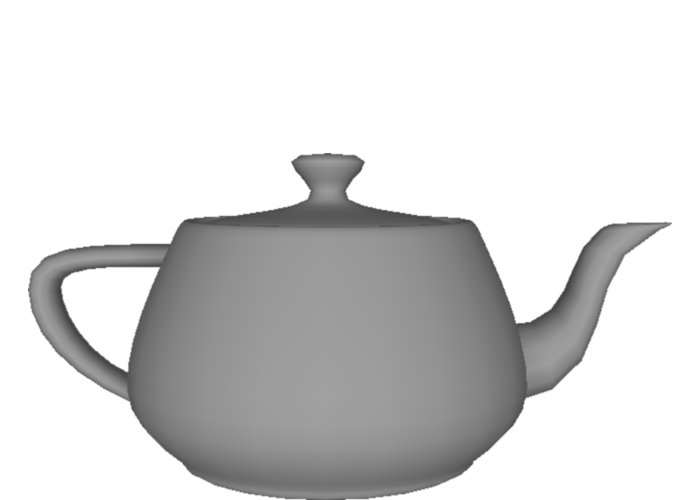}} \hfill
	\captionsetup[subfigure]{labelformat=empty}
	\subfloat[Stanford Bunny]{\includegraphics[width=.32\linewidth]{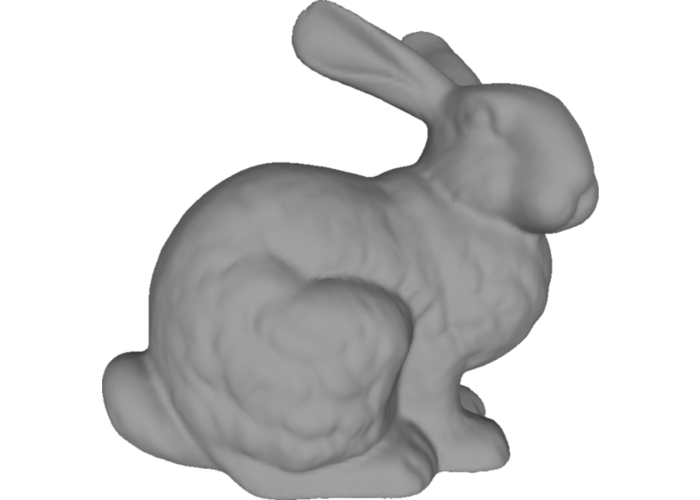}} \hfill
	\subfloat[Stanford Dragon]{\includegraphics[width=.32\linewidth]{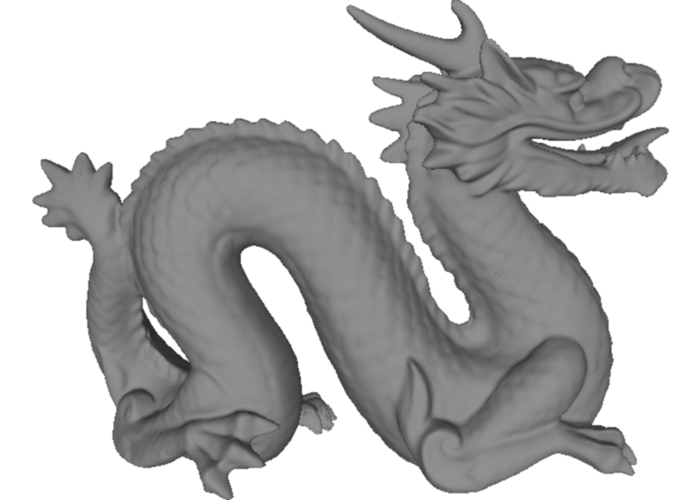}} \hfill
	\vspace{-1ex}
	\captionsetup[subfigure]{}
	\subfloat[]{\includegraphics[width=.32\linewidth]{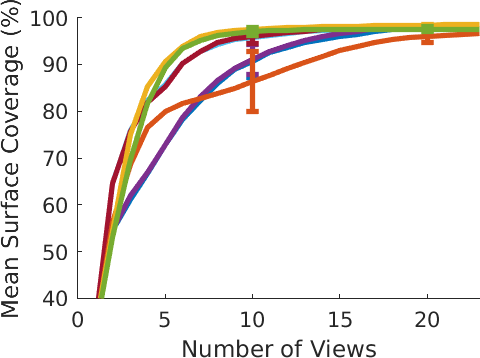}} \hfill
	\captionsetup[subfigure]{labelformat=empty}
	\subfloat[]{\includegraphics[width=.32\linewidth]{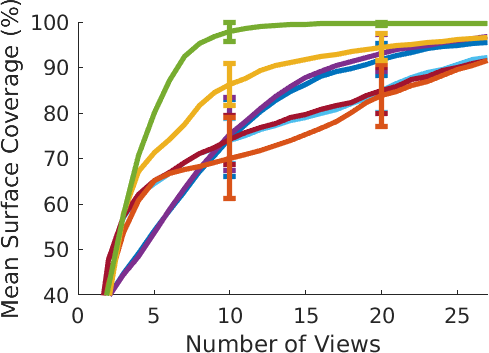}} \hfill
	\subfloat[]{\includegraphics[width=.32\linewidth]{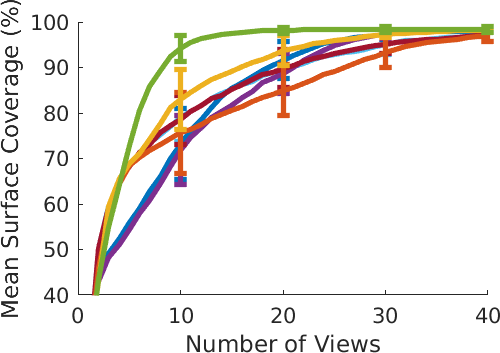}} \hfill
	\vspace{-3ex}
	\captionsetup[subfigure]{}
	\subfloat[]{\includegraphics[width=.32\linewidth]{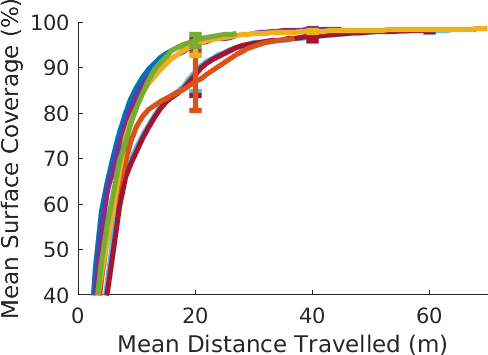}} \hfill
	\captionsetup[subfigure]{labelformat=empty}
	\subfloat[]{\includegraphics[width=.32\linewidth]{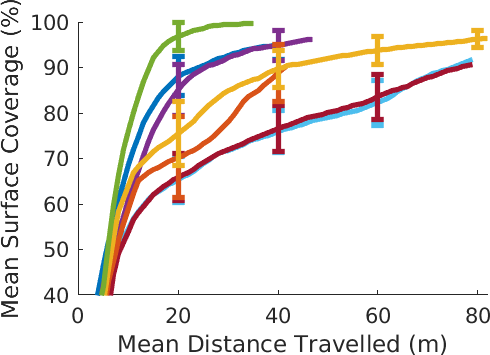}} \hfill
	\subfloat[]{\includegraphics[width=.32\linewidth]{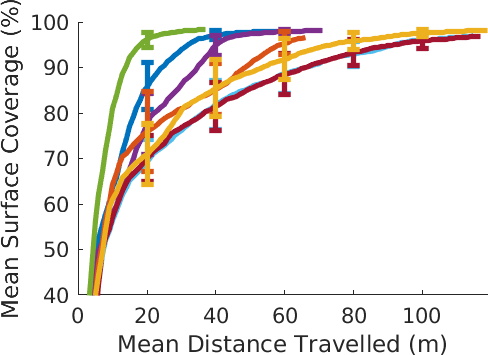}} \hfill
	\vspace{-3ex}
	\captionsetup[subfigure]{}
	\subfloat[]{\includegraphics[width=.32\linewidth]{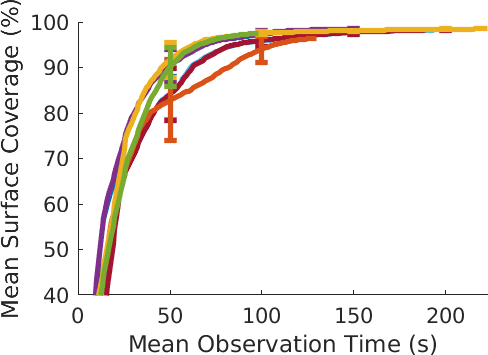}} \hfill
	\captionsetup[subfigure]{labelformat=empty}
	\subfloat[]{\includegraphics[width=.32\linewidth]{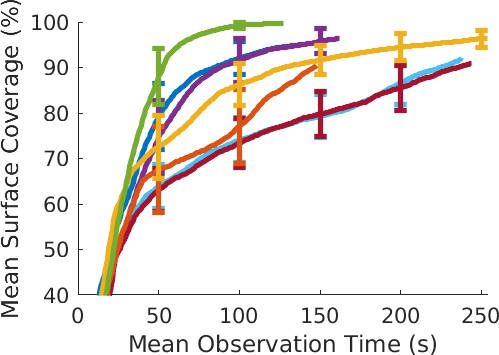}} \hfill
	\subfloat[]{\includegraphics[width=.32\linewidth]{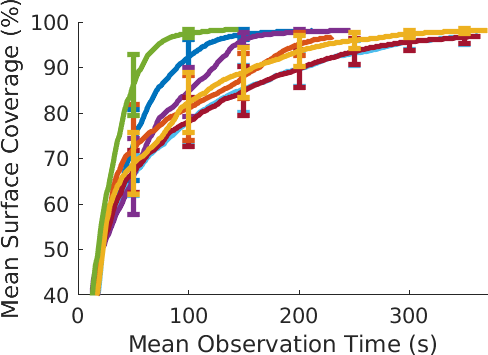}} \hfill
	\vspace{-3ex}
	\subfloat[]{\includegraphics[width=0.96\linewidth]{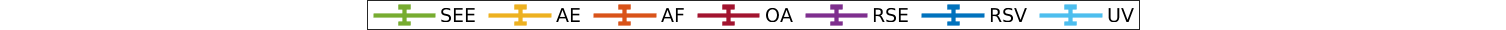}} \\
	\vspace{-3ex}
	\subfloat[]{
		\begin{adjustbox}{width=\linewidth,center}
			\begin{tabular}{ccccccccccccc}
								\multicolumn{1}{c}{} & \multicolumn{3}{c}{Newell Teapot (11 / 17 / 23 Views)} & \multicolumn{3}{c}{Stanford Bunny (13 / 20 / 27 Views)}  & \multicolumn{3}{c}{Stanford Dragon (20 / 30 / 40 Views)} \\
				\cmidrule(lr){2-4}
				\cmidrule(lr){5-7}
				\cmidrule(lr){8-10}
				\multicolumn{1}{c}{} & Coverage (\%) & Distance (m) & Time (s) & Coverage (\%) & Distance (m) & Time (s) & Coverage (\%) & Distance (m) & Time (s) \\ 
				\midrule
				SEE  & 97.2 / 97.6 / 97.6 & 25 / 28 / \textbf{28}& 89 / 98 / \textbf{99} & \textbf{99.3} / \textbf{99.7} / \textbf{99.7} & 28 / 35 / \textbf{36} & 101 / 124 / \textbf{127} & \textbf{98.2} / \textbf{98.3} / \textbf{98.4}& 34 / 37 / \textbf{38} & 129 / \textbf{144} / \textbf{145}  \\
				AE  & \textbf{97.8} / \textbf{98.3} / \textbf{98.5} & 36 / 55 / 71 & 113 / 175 / 224 & 90.4 / 94.4 / 96.6 & 42 / 65 / 83 & 132 / 200 / 255 & 93.7 / 97.3 / 98.2 & 66 / 95 / 119 & 201 / 295 / 372  \\
				AF  & 87.6 / 94.8 / 96.6 & 21 / 30 / 38 & 72 / 105 / 130 & 72.8 / 83.8 / 91.5 & 25 / 34 / 42 & 84 / 119 / 149 & 85.0 / 93.3 / 96.9 & 39 / 54 / 67 & 131 / 183 / 230  \\
				OA  & 96.5 / 98.0 / 98.4 & 37 / 53 / 66 & 111 / 164 / 206 & 77.8 / 85.0 / 91.4 & 43 / 63 / 80 & 132 / 194 / 245 & 89.8 / 95.1 / 97.1 & 65 / 92 / 118 & 201 / 284 / 365  \\
				RSE  & 92.7 / 97.6 / \textbf{98.5} & 17 / 30 / 46 & 58 / 101 / 151 & 83.6 / 93.2 / 96.8 & 19 / 33 / 48 & 66 / 112 / 162 & 88.6 / 97.2 / 98.2 & 33 / 48 / 72 & 117 / 170 / 247  \\
				RSV  & 92.5 / 97.0 / 98.4 & \textbf{14} / \textbf{27} / 41 & \textbf{53} / \textbf{94} / 138 & 82.7 / 91.8 / 95.6 & \textbf{16} / \textbf{28} / 42 & \textbf{56} / \textbf{97} / 143 & 91.6 / 97.3 / 98.1 & \textbf{26} / \textbf{41} / 60 & \textbf{97} / 151 / 212  \\
				UV  & 96.3 / 97.9 / 98.4 & 37 / 53 / 68 & 110 / 162 / 207 & 77.2 / 84.9 / 92.1 & 44 / 64 / 79 & 128 / 191 / 238 & 89.6 / 95.1 / 97.1 & 64 / 92 / 116 & 199 / 288 / 363  \\
				\bottomrule
			\end{tabular}
	\end{adjustbox}}
	\vspace{-3ex}
	\caption{A comparison of \gls{see} and the evaluated volumetric approaches in the UR10 simulation environment for $100$ experiments on the Newell Teapot, Stanford Bunny and Stanford Dragon. The graphs show, from top to bottom, the mean surface coverage relative to the number of views, the mean surface coverage relative to travel distance and the mean surface coverage relative to observation time. The mean surface coverage axes start at $40$\% to highlight the algorithm performance at completion. The error bars denote one standard deviation around the mean. The table shows each algorithm's mean surface coverage, distance travelled and observation time over all $100$ experiments on each model for $50$\%, $75$\% and $100$\% of the maximum views taken respectively, with the best value highlighted in bold.}
	\figlabel{see-results-1}
\end{figure*}

\begin{figure*}[tpb]
	\centering
	\captionsetup[subfigure]{}
	\captionsetup[subfigure]{labelformat=empty}
	\captionsetup[subfigure]{justification=centering}
	
	\subfloat[Stanford Armadillo]{\includegraphics[width=.32\linewidth]{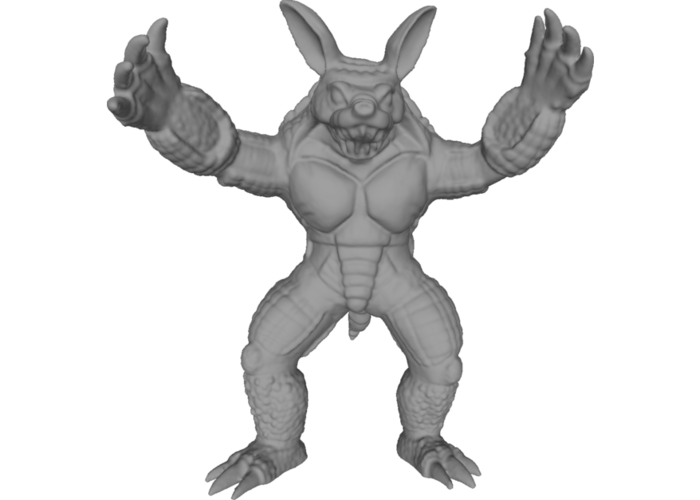}} \hfill
	\captionsetup[subfigure]{labelformat=empty}
	\subfloat[Happy Buddha]{\includegraphics[width=.32\linewidth]{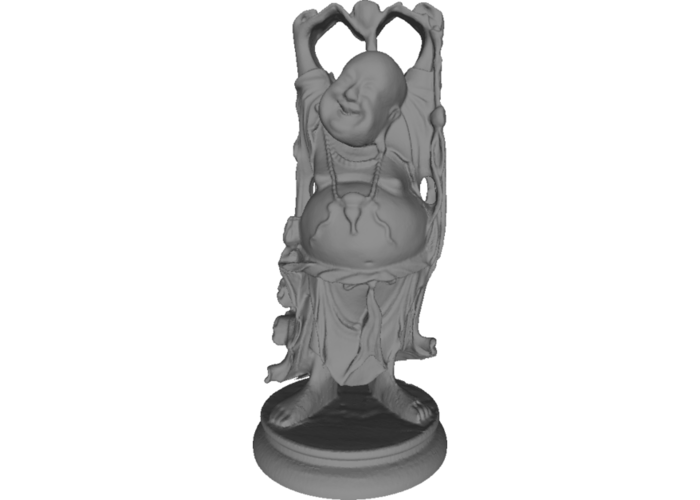}} \hfill
	\subfloat[Helix]{\includegraphics[width=.32\linewidth]{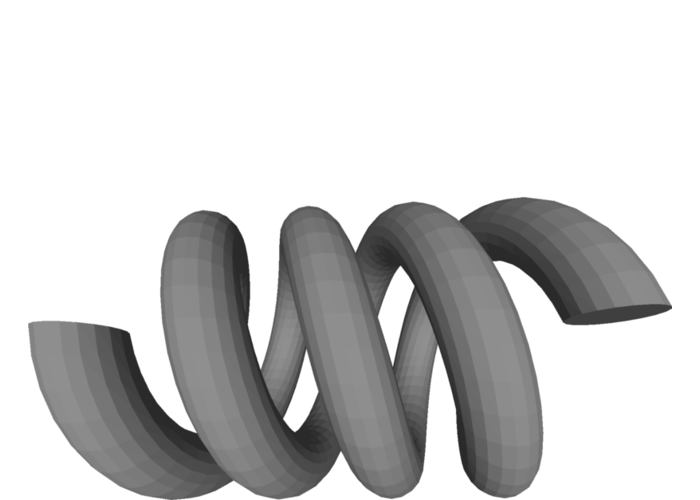}} \hfill
	\vspace{-1ex}
	\captionsetup[subfigure]{}
	\subfloat[]{\includegraphics[width=.32\linewidth]{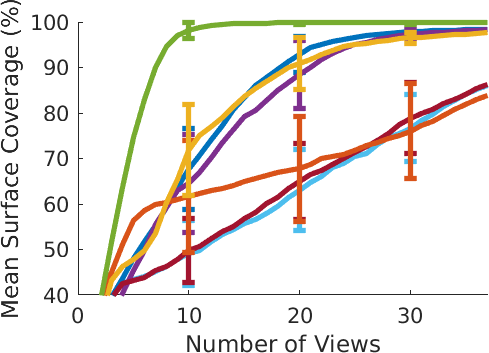}} \hfill
	\captionsetup[subfigure]{labelformat=empty}
	\subfloat[]{\includegraphics[width=.32\linewidth]{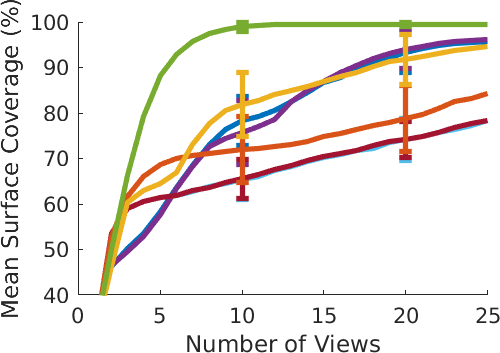}} \hfill
	\subfloat[]{\includegraphics[width=.32\linewidth]{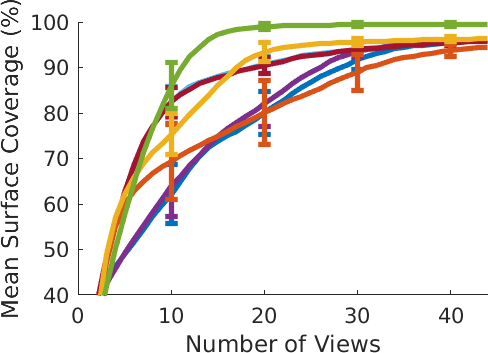}} \hfill
	\vspace{-3ex}
	\captionsetup[subfigure]{}
	\subfloat[]{\includegraphics[width=.32\linewidth]{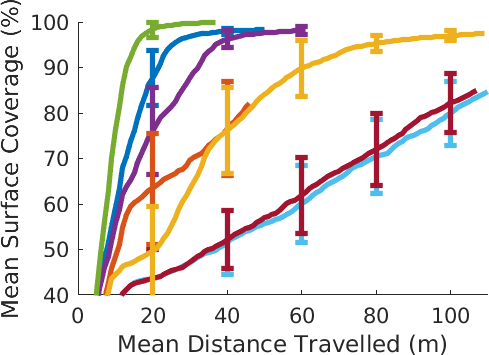}} \hfill
	\captionsetup[subfigure]{labelformat=empty}
	\subfloat[]{\includegraphics[width=.32\linewidth]{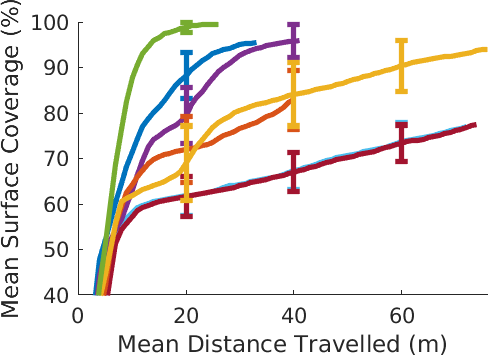}} \hfill
	\subfloat[]{\includegraphics[width=.32\linewidth]{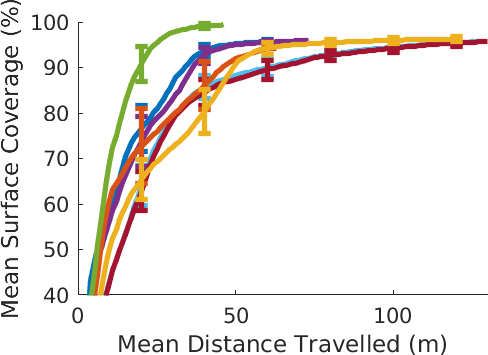}} \hfill
	\vspace{-3ex}
	\captionsetup[subfigure]{}
	\subfloat[]{\includegraphics[width=.32\linewidth]{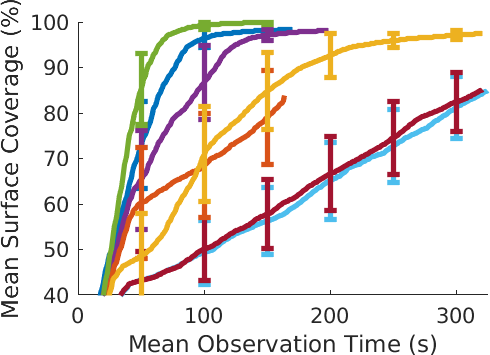}} \hfill
	\captionsetup[subfigure]{labelformat=empty}
	\subfloat[]{\includegraphics[width=.32\linewidth]{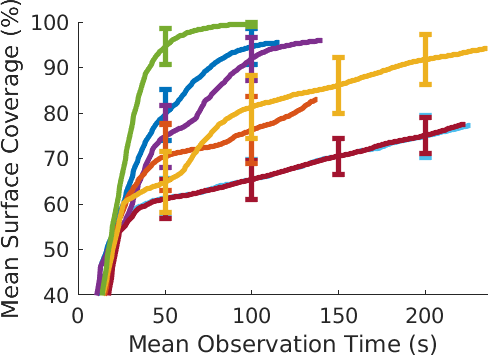}} \hfill
	\subfloat[]{\includegraphics[width=.32\linewidth]{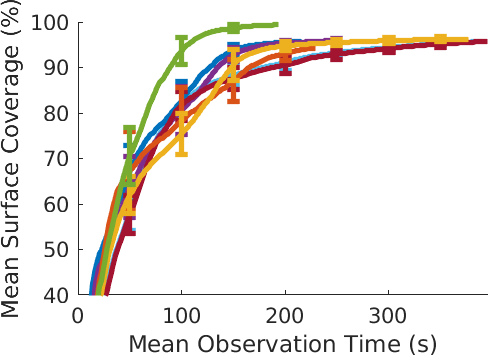}} \hfill
	\vspace{-3ex}
	\subfloat[]{\includegraphics[width=0.96\linewidth]{results/small_series_legend}}\\
	\vspace{-3ex}
	\subfloat[]{
		\begin{adjustbox}{width=\linewidth,center}
			\begin{tabular}{ccccccccccccc}
				\multicolumn{1}{c}{} & \multicolumn{3}{c}{Stanford Armadillo (18 / 28 / 37 Views)} & \multicolumn{3}{c}{Happy Buddha (12 / 19 / 25 Views)} & \multicolumn{3}{c}{Helix (22 / 33 / 44 Views)} \\
				\cmidrule(lr){2-4}
				\cmidrule(lr){5-7}
				\cmidrule(lr){8-10}
				\multicolumn{1}{c}{} & Coverage (\%) & Distance (m) & Time (s) & Coverage (\%) & Distance (m) & Time (s) & Coverage (\%) & Distance (m) & Time (s) \\ 
				\midrule
				SEE  & \textbf{99.8} / \textbf{99.8} / \textbf{99.9} & 33 / 38 / \textbf{38} & 134 / 149 / \textbf{150} & \textbf{99.4} / \textbf{99.5} / \textbf{99.6} & 23 / 26 / \textbf{26} & 89 / 100 / \textbf{101} & \textbf{99.2} / \textbf{99.4} / \textbf{99.4} & 42 / 47 / \textbf{47} & 173 / 193 / \textbf{194}  \\
				AE  & 88.4 / 96.0 / 97.6 & 57 / 86 / 110 & 169 / 253 / 322 & 84.1 / 91.3 / 94.7 & 40 / 63 / 77 & 125 / 193 / 237 & 94.4 / 95.8 / 96.3 & 61 / 93 / 121 & 187 / 289 / 378  \\
				AF  & 66.8 / 74.5 / 83.8 & 26 / 37 / 47 & 89 / 130 / 164 & 72.6 / 77.8 / 84.4 & 23 / 33 / 40 & 77 / 113 / 139 & 82.1 / 91.1 / 94.4 & 34 / 48 / 63 & 120 / 174 / 230  \\
				OA  & 61.3 / 75.5 / 86.2 & 59 / 87 / 108 & 171 / 255 / 322 & 67.5 / 73.7 / 78.4 & 42 / 61 / 74 & 121 / 182 / 224 & 91.4 / 94.5 / 95.8 & 71 / 103 / 130 & 216 / 312 / 397  \\
				RSE  & 85.0 / 96.7 / 98.3 & 28 / 42 / 61 & 95 / 141 / 199 & 78.7 / 93.1 / 96.1 & 20 / 31 / 42 & 67 / 105 / 142 & 84.9 / 94.6 / 96.0 & 32 / 47 / 73 & 118 / 172 / 254  \\
				RSV  & 89.5 / 97.5 / 98.4 & \textbf{21} / \textbf{34} / 51 & \textbf{75} / \textbf{118} / 170 & 80.7 / 92.2 / 95.7 & \textbf{14} / \textbf{24} / 34 & \textbf{52} / \textbf{84} / 117 & 82.8 / 93.6 / 95.8 & \textbf{27} / \textbf{40} / 61 & \textbf{100} / \textbf{148} / 216  \\
				UV  & 59.5 / 74.6 / 86.1 & 59 / 88 / 111 & 172 / 260 / 326 & 67.3 / 73.5 / 78.4 & 40 / 59 / 73 & 118 / 181 / 226 & 91.6 / 94.5 / 95.8 & 70 / 100 / 127 & 209 / 304 / 391  \\
				\bottomrule
			\end{tabular}
	\end{adjustbox}}
	\vspace{-3ex}
	\caption{A comparison of \gls{see} and the evaluated volumetric approaches in the UR10 simulation environment for $100$ experiments on the Stanford Armadillo, Happy Buddha and Helix. The graphs show, from top to bottom, the mean surface coverage relative to the number of views, the mean surface coverage relative to travel distance and the mean surface coverage relative to observation time. The mean surface coverage axes start at $40$\% to highlight the algorithm performance at completion. The error bars denote one standard deviation around the mean. The table shows each algorithm's mean surface coverage, distance travelled and observation time over all $100$ experiments on each model for $50$\%, $75$\% and $100$\% of the maximum views taken respectively, with the best value highlighted in bold.}
	\figlabel{see-results-2}
\end{figure*}

\begin{figure*}[tpb]
	\centering
	\captionsetup[subfigure]{}
	\captionsetup[subfigure]{labelformat=empty}
	\captionsetup[subfigure]{justification=centering}
	
	\subfloat[Statue of Liberty]{\includegraphics[width=.32\linewidth]{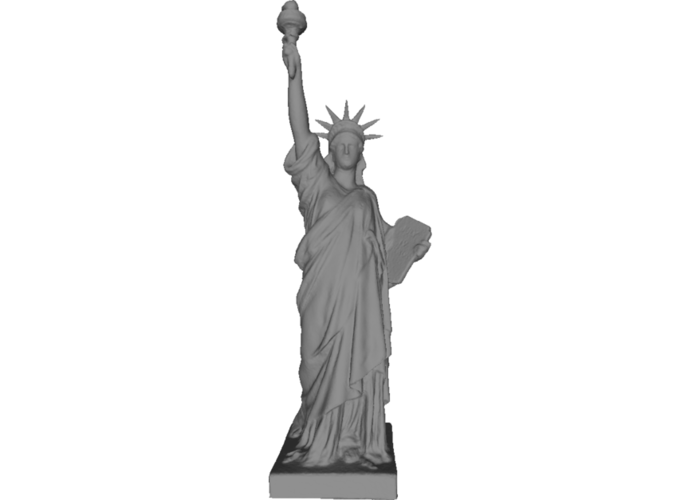}} \hfill
	\captionsetup[subfigure]{labelformat=empty}
	\subfloat[Radcliffe Camera]{\includegraphics[width=.32\linewidth]{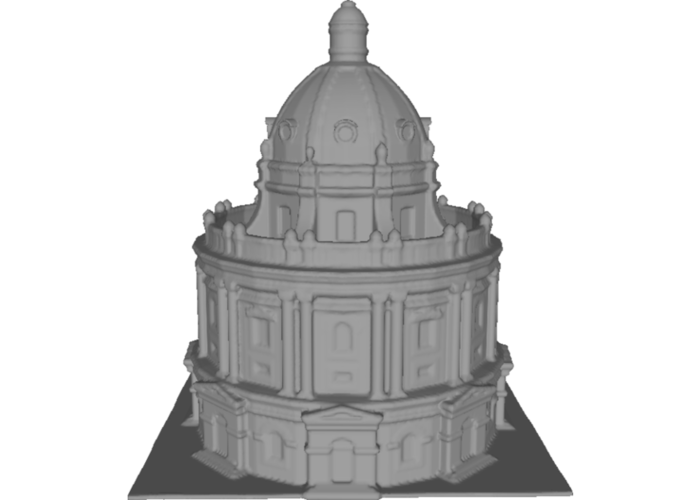}} \hfill
	\subfloat[Notre-Dame de Paris]{\includegraphics[width=.32\linewidth]{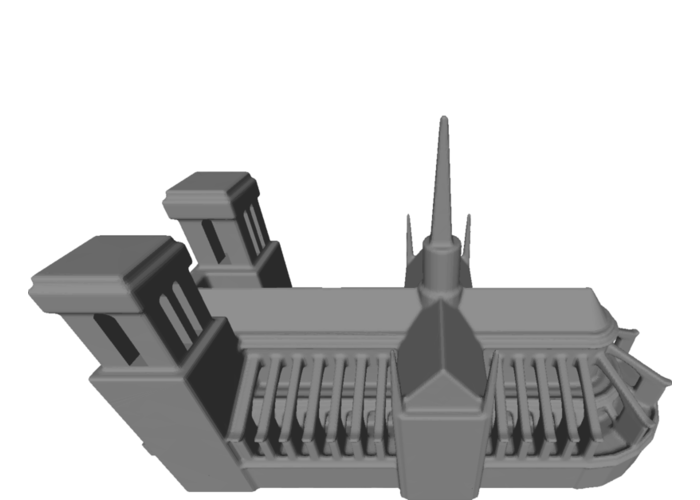}} \hfill
	\vspace{-1ex}
	\captionsetup[subfigure]{}
	\subfloat[]{\includegraphics[width=.32\linewidth]{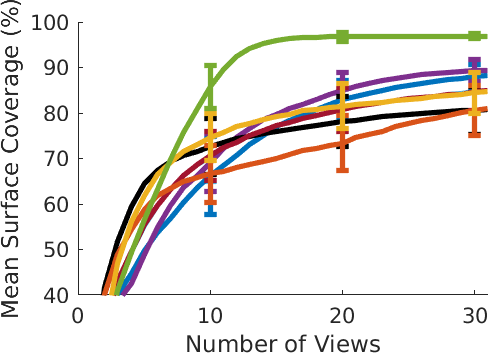}} \hfill
	\captionsetup[subfigure]{labelformat=empty}
	\subfloat[]{\includegraphics[width=.32\linewidth]{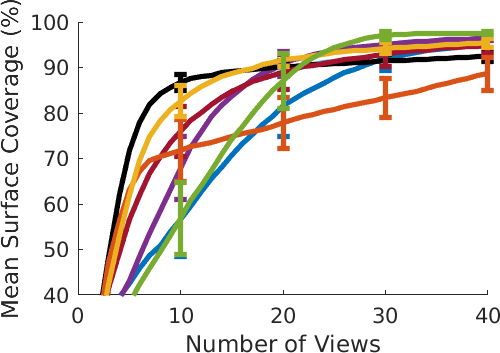}} \hfill
	\subfloat[]{\includegraphics[width=.32\linewidth]{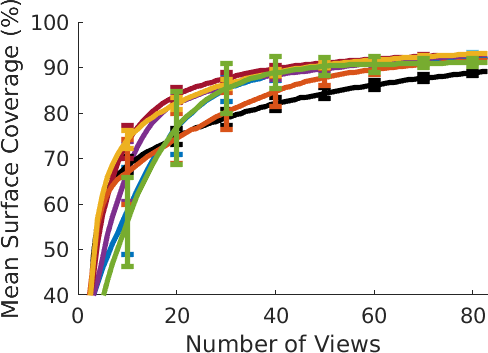}} \hfill
	\vspace{-3ex}
	\captionsetup[subfigure]{}
	\subfloat[]{\includegraphics[width=.32\linewidth]{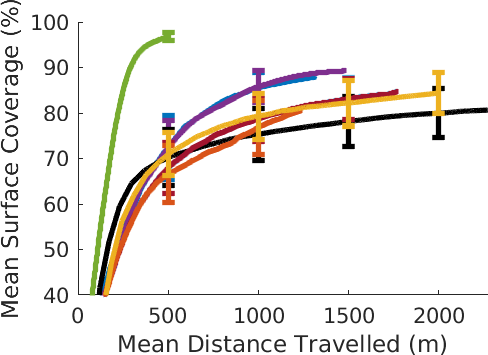}} \hfill
	\captionsetup[subfigure]{labelformat=empty}
	\subfloat[]{\includegraphics[width=.32\linewidth]{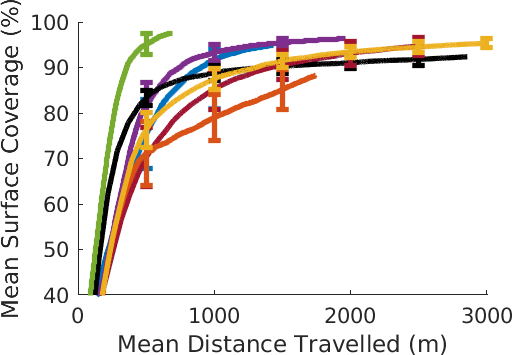}} \hfill
	\subfloat[]{\includegraphics[width=.32\linewidth]{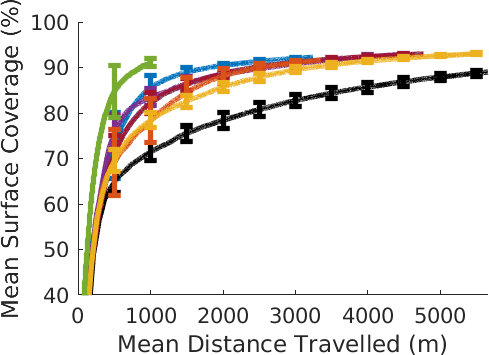}} \hfill
	\vspace{-3ex}
	\captionsetup[subfigure]{}
	\subfloat[]{\includegraphics[width=.32\linewidth]{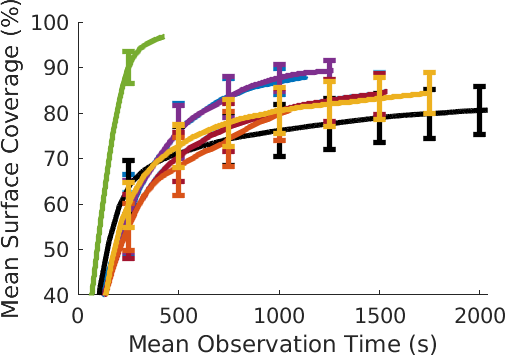}} \hfill
	\captionsetup[subfigure]{labelformat=empty}
	\subfloat[]{\includegraphics[width=.32\linewidth]{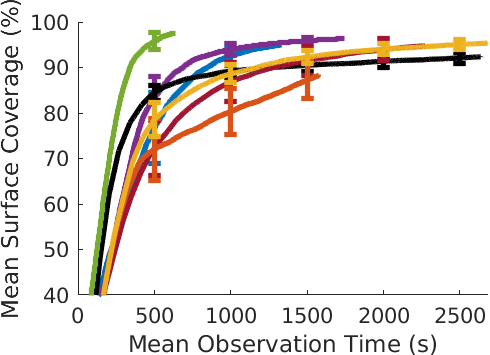}} \hfill
	\subfloat[]{\includegraphics[width=.32\linewidth]{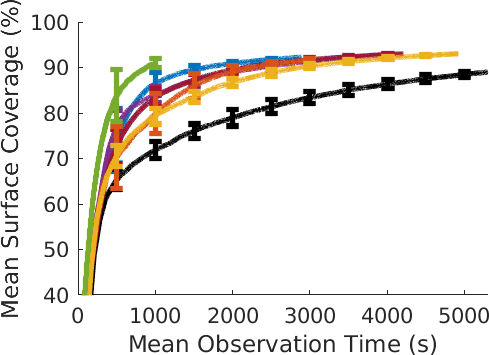}} \hfill
	\vspace{-3ex}
	\subfloat[]{\includegraphics[width=0.96\linewidth]{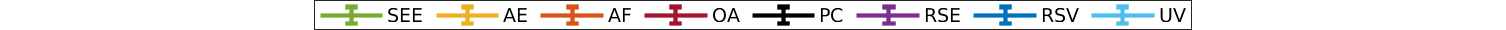}} \\
	\vspace{-3ex}
	\subfloat[]{
		\begin{adjustbox}{width=0.995\linewidth,center}
			\begin{tabular}{ccccccccccccc}
				\multicolumn{1}{c}{} & \multicolumn{3}{c}{Statue of Liberty (15 / 23 / 31 Views)} & \multicolumn{3}{c}{Radcliffe Camera (20 / 30 / 40 Views)}  & \multicolumn{3}{c}{Notre-Dame de Paris (41 / 62 / 83 Views)} \\
				\cmidrule(lr){2-4}
				\cmidrule(lr){5-7}
				\cmidrule(lr){8-10}
				\multicolumn{1}{c}{} & Coverage (\%) & Distance (m) & Time (s) & Coverage (\%) & Distance (m) & Time (s) & Coverage (\%) & Distance (m) & Time (s) \\ 
				\midrule
				SEE  & \textbf{95.9} / \textbf{96.9} / \textbf{96.9} & \textbf{436} / \textbf{501} / \textbf{504} & \textbf{373} / \textbf{428} / \textbf{431} & 87.1 / \textbf{97.0} / \textbf{97.6} & \textbf{323} / \textbf{622} / \textbf{692} & \textbf{310} / \textbf{574} / \textbf{636} & 89.0 / 90.8 / 91.0 & \textbf{760} / \textbf{969} / \textbf{1000} & \textbf{786} / \textbf{980} / \textbf{1008}  \\
				AE  & 79.2 / 82.4 / 84.7 & 988 / 1541 / 2007 & 861 / 1348 / 1754 & \textbf{91.7} / 94.2 / 95.4 & 1522 / 2317 / 3009 & 1362 / 2065 / 2683 & 89.1 / 91.8 / \textbf{93.1} & 2884 / 4271 / 5542 & 2585 / 3810 / 4922  \\
				AF  & 70.1 / 76.0 / 81.1 & 663 / 964 / 1242 & 572 / 822 / 1053 & 77.8 / 83.3 / 88.6 & 924 / 1349 / 1742 & 840 / 1220 / 1577 & 85.1 / 89.8 / 91.6 & 1562 / 2457 / 3350 & 1410 / 2214 / 3008  \\
				OA  & 77.2 / 82.3 / 85.0 & 882 / 1365 / 1778 & 768 / 1187 / 1541 & 89.0 / 93.0 / 94.9 & 1297 / 1953 / 2546 & 1172 / 1752 / 2275 & \textbf{89.9} / \textbf{92.0} / 93.0 & 2393 / 3614 / 4769 & 2127 / 3197 / 4205  \\
				PC  & 75.9 / 79.2 / 80.7 & 1087 / 1718 / 2272 & 965 / 1538 / 2040 & 90.1 / 91.5 / 92.5 & 1441 / 2204 / 2859 & 1336 / 2038 / 2638 & 82.3 / 86.5 / 89.1 & 2887 / 4327 / 5659 & 2713 / 4053 / 5302  \\
				RSE  & 79.8 / 87.1 / 89.4 & 697 / 1107 / 1487 & 600 / 948 / 1272 & 90.8 / 95.1 / 96.4 & 777 / 1377 / 1967 & 714 / 1237 / 1744 & 88.5 / 91.2 / 92.9 & 2033 / 3184 / 4422 & 1842 / 2863 / 3945  \\
				RSV  & 77.1 / 85.0 / 88.2 & 593 / 957 / 1315 & 517 / 828 / 1136 & 81.5 / 91.9 / 95.1 & 609 / 1012 / 1438 & 580 / 944 / 1323 & 88.7 / 91.1 / 92.3 & 1426 / 2343 / 3244 & 1301 / 2107 / 2891  \\
				UV  & 77.2 / 82.3 / 85.0 & 882 / 1365 / 1776 & 768 / 1187 / 1538 & 89.0 / 93.0 / 94.9 & 1295 / 1952 / 2543 & 1171 / 1752 / 2272 & \textbf{89.9} / \textbf{92.0} / 93.0 & 2397 / 3618 / 4772 & 2134 / 3205 / 4214  \\
				\bottomrule
			\end{tabular}
	\end{adjustbox}}
	\vspace{-3ex}
	\caption{A comparison of \gls{see} and the evaluated volumetric approaches in the UAV simulation environment for $100$ experiments on the Statue of Liberty, Radcliffe Camera and Notre-Dame de Paris. The graphs show, from top to bottom, the mean surface coverage relative to the number of views, the mean surface coverage relative to travel distance and the mean surface coverage relative to observation time. The mean surface coverage axes start at $40$\% to highlight the algorithm performance at completion. The error bars denote one standard deviation around the mean. The table shows each algorithm's mean surface coverage, distance travelled and observation time over all $100$ experiments on each model for $50$\%, $75$\% and $100$\% of the maximum views taken respectively, with the best value highlighted in bold.}
	\figlabel{see-results-3}
\end{figure*}

\subsection{Discussion}
\seclabel{disc}

The results (Figs. \ref{fig:see-results-1}--\ref{fig:see-results-3}) show that a measurement-direct \gls{nbv} approach with a pointcloud representation captures highly complete scene observations with greater efficiency than structured volumetric approaches. \gls{see} achieves similar or better surface coverage than the volumetric approaches on every tested model while using shorter travel distances and less observation time. For the maximum number of views taken, \gls{see} captures equivalent or greater surface coverage per view, unit of travel distance and unit of observation time than the volumetric approaches on every model. 

When considering $50$\% and $75$\% of the number of views taken, \gls{see} captures equivalent or greater surface coverage per view than all of the volumetric approaches on every model except one. The surface coverage obtained per unit of distance and observation time is greater than all of the volumetric approaches on the large models and most of them on the small models. The following sections discuss the performance of \gls{see} on individual small \psecref{disc-small} and large \psecref{disc-large} models, and review limitations of the volumetric approaches \psecref{disc-vol}.  

\subsubsection{Small Models}
\seclabel{disc-small}

\gls{see} captures similar or better surface coverage than all of the volumetric approaches on every small model using less travel distance and a shorter observation time. It also attains equivalent or greater surface coverage per view than all of the volumetric approaches on every model for $50$\%, $75$\% and $100$\% of the maximum number of views taken.

\gls{see} obtains higher surface coverage than all of the volumetric approaches on every small model except for the Newell Teapot. It achieves marginally lower surface coverage on the Newell Teapot but requires a shorter travel distance and less observation time to do so. This is because sensor noise causes the intersections between the teapot body and the handle to be prematurely classified as fully observed.

\gls{see} achieves greater surface coverage per unit of travel distance and observation time than all of the volumetric approaches on every small model when considering the maximum number of views taken. For $75$\% of the maximum views taken, \gls{see} achieves equivalent or greater surface coverage per unit of travel distance and observation time than all but one volumetric approach on every model except the Stanford Bunny and Stanford Armadillo, where RSV achieves a greater surface coverage per unit of travel distance. For $50$\% of the maximum views taken, \gls{see} achieves equivalent or greater surface coverage per unit of travel distance and observation time than most of the volumetric approaches on every model, except for RSE and RSV which achieve greater coverage relative to both metrics.  

This demonstrates that these volumetric approaches are initially able to attain greater surface coverage in less time and distance than \gls{see} but that their progress slows as they approach a complete observation since they require more time and distance to capture measurements from surfaces that are challenging to observe (e.g., between the folds of the Dragon and inside the Helix), resulting in a worse overall observation efficiency than \gls{see}.

\subsubsection{Large Models}
\seclabel{disc-large}

\gls{see} obtains similar or better surface coverage than all of the evaluated volumetric approaches on every large model using less travel distance and observation time. It also attains equivalent or greater surface coverage per unit of travel distance and unit of observation time than all of the volumetric approaches on every model, for $50$\%, $75$\% and $100$\% of the maximum views taken.

\gls{see} obtains equivalent or greater surface coverage per view than every volumetric approach on every model for $75$\% and $100$\% of the maximum views taken. For $50$\% of the maximum views taken it attains equivalent or greater surface coverage per view than every volumetric approach on every model except for the Radcliffe Camera, where \gls{see} captures slightly less coverage per view than all of the volumetric approaches except for AF and RSV. This demonstrates that while some of the volumetric approaches are initially able to observe more of the model with fewer views their progress again slows as they approach a complete observation since they struggle to capture some challenging surfaces (e.g., inside the balconies) and are less efficient than \gls{see} overall.     

\gls{see} achieves higher surface coverage than all of the volumetric approaches on every large model except for the Notre-Dame de Paris, where it performs marginally worse than all of the volumetric approaches except PC. This is because it took \gls{see} more observation time to plan and capture views of the flying buttresses without successfully increasing surface coverage underneath these uniquely challenging features; however, \gls{see} still observes the Notre-Dame de Paris using a shorter travel distance and a lower observation time than all of the volumetric approaches.

\subsubsection{Volumetric Limitations}
\seclabel{disc-vol}

The results demonstrate that the final surface coverage of the volumetric approaches depends upon the surface area of a model relative to the volume of its bounding box. This variation is most noticeable for the AF, OA, UV and PC approaches, which perform significantly worse on models with high surface-to-volume ratios, either locally (e.g., Stanford Bunny ears and Stanford Armadillo limbs) or globally (e.g., the Happy Buddha and Statue of Liberty). These algorithms prioritise voxels that are visible from a previous view, which limits the final surface coverage of objects not easily observed with overlapping views. The PC algorithm is the most affected as it only prioritizes occluded voxels. This made a fair evaluation on the small models unfeasible due to the small viewing frustum of the simulated sensor. 

SEE consistently observes all of the models more efficiently than the evaluated volumetric approaches. It obtains equivalent or better observations of every model using shorter travel distances and lower observation times. \gls{see} achieves this by proposing and selecting views that typically obtain greater improvements in surface coverage per unit of distance travelled, particularly for surfaces that are challenging to observe. The surface coverage obtained per unit of observation time is also frequently higher due to the proactive handling of known occlusions.

\section{Real-World Experiments}
\seclabel{real-exp}

\begin{figure*}[tpb]
	\centering
	\captionsetup[subfigure]{}
	\captionsetup[subfigure]{labelformat=empty}
	\captionsetup[subfigure]{justification=centering}
	
	\subfloat[Photograph of the Oxford Deer]{\includegraphics[width=.32\linewidth]{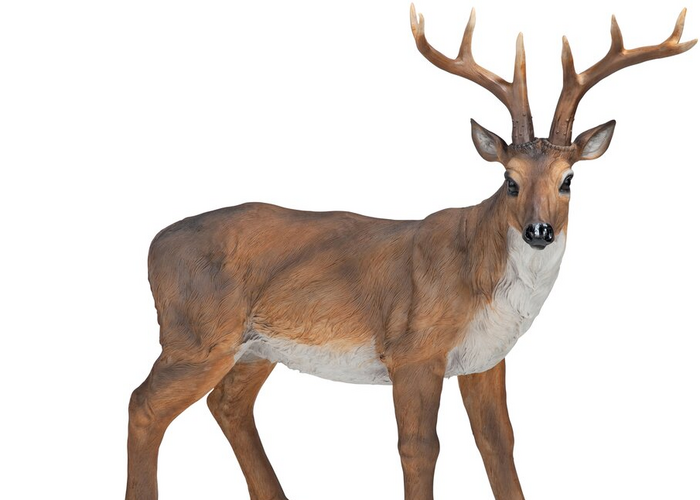}} \hfill
	\subfloat[Coloured Pointcloud]{\includegraphics[width=.32\linewidth]{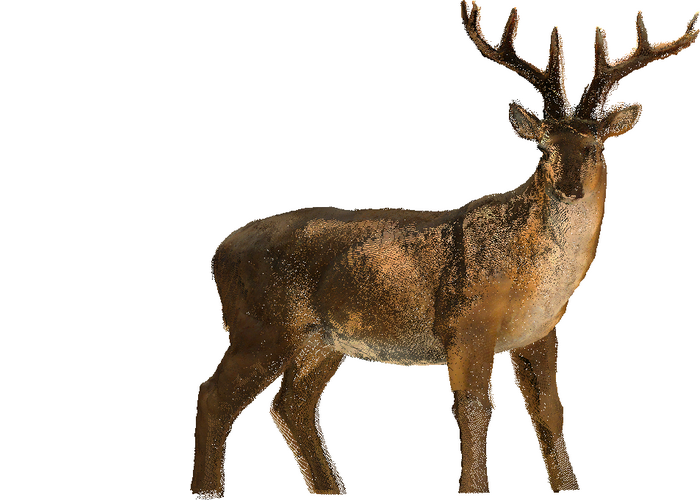}} \hfill
	\subfloat[Generated Greyscale Mesh]{\includegraphics[width=.32\linewidth]{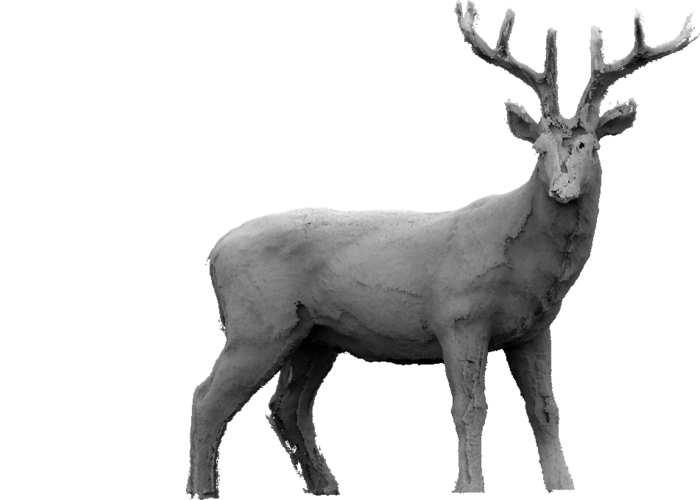}} \hfill
	\vspace{-1ex}
	\captionsetup[subfigure]{}
	\subfloat[]{\includegraphics[width=.32\linewidth]{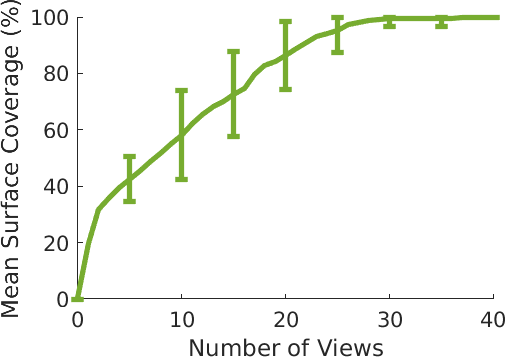}} \hfill
	\subfloat[]{\includegraphics[width=.32\linewidth]{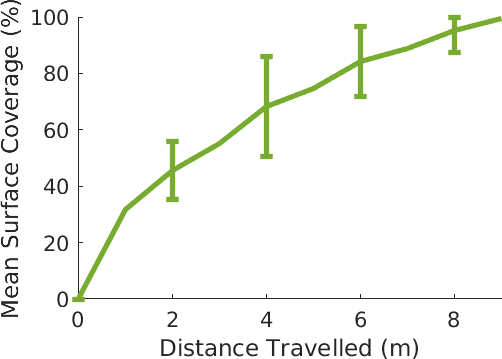}} \hfill
	\subfloat[]{\includegraphics[width=.32\linewidth]{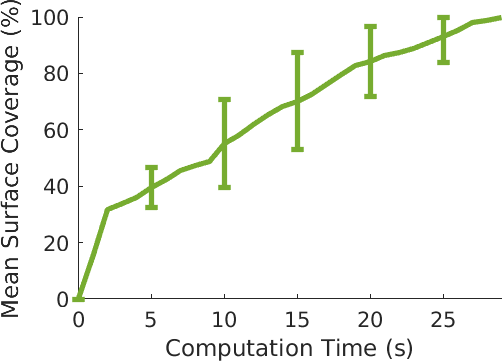}} \hfill
	\vspace{-3ex}
	\subfloat[]{\includegraphics[width=0.96\linewidth]{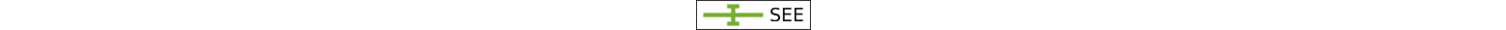}} \\
	\vspace{-3ex}
	\subfloat[]{
		\begin{adjustbox}{width=\linewidth,center}
			\begin{tabular}{cccccccc}
				\cmidrule(lr){4-5}
				\cmidrule(lr){6-7}
				\multicolumn{3}{c}{} & \multicolumn{2}{c}{Computation Time (s)} & \multicolumn{2}{c}{Execution Time (s)}  & \multicolumn{1}{c}{} \\	
				\cmidrule(lr){2-8}
				\multicolumn{1}{c}{} & Views & Distance Travelled (m) & NBV Planning Time (s) & Path Planning Time (s) & Movement Time (s) & Measurement Time (s) & Total Observation Time (s) \\ 
				\midrule
				SEE  & 27 & 9 & 19 & 9  & 83 & 344 & 455 \\
				\bottomrule
			\end{tabular}
	\end{adjustbox}}
	\caption{A demonstration of \gls{see} on the UR10 platform for $20$ independent experiments observing the Oxford Deer. The images show, from left to right, a photograph of the Oxford Deer, the coloured pointcloud obtained by \gls{see} for a representative run and a greyscale mesh generated from that pointcloud using Open3D, respectively. The graphs show, from left to right, the mean surface coverage relative to the number of views, distance travelled and computation time, respectively. The error bars denote one standard deviation around the mean. The surface coverage is computed relative to the final pointcloud since no ground truth is available. The table shows the final mean number of views captured, distance travelled, \gls{nbv} planning time, path planning time, movement time, measurement time and total observation time over all $20$ experiments.}
	\figlabel{real-results}
\end{figure*}   

\begin{table}
	\caption{The field-of-view in degrees, $\theta_\mathrm{x}$ and $\theta_\mathrm{y}$, and resolution in pixels, $w_\mathrm{x}$ and $w_\mathrm{y}$, of the Intel RealSense L515 used in the experiments in \tsecref{real-exp}.}
	\centering
	\begin{tabular}{@{}lll@{}}
		\toprule
		& Intel RealSense L515 & Units \\ \midrule
		$\theta_\mathrm{x}$ & $70$ & degrees \\
		$\theta_\mathrm{y}$ & $43$ & degrees \\
		$w_\mathrm{x}$ & $640$ & pixels \\
		$w_\mathrm{y}$ & $480$ & pixels \\ \bottomrule
	\end{tabular}
	\tbllabel{real-sensor-table}
\end{table}  

\gls{see} was deployed on a UR10 robotic arm to evaluate its real-world performance by observing a deer statue, the \emph{Oxford Deer}, for $20$ independent experiments. The statistically significant results demonstrate that \gls{see} performs well observing an object with varied texture, geometry and self-occlusions using a real platform. It is able to capture high-quality observations of the Oxford Deer despite the complex measurement noise associated with a real sensor, which can vary with the view orientation, surface geometry and texture.

The UR10 platform \figref{inspector-real} mirrors the UR10 simulation environment, with objects on a turntable being captured by an Intel RealSense L515 \ptblref{real-sensor-table} affixed to the UR10 end effector. The turntable and UR10 are jointly controlled to obtain a complete observation of the unknown object. 

The observation pipeline was extended to handle the noise of a real-world system.
Collision-free paths between views were planned with \gls{aitstar} and executed by MoveIt. The sensor pose was then held steady for $5$ seconds before capturing measurements to ensure stability. A radius-based outlier filter was applied to captured measurements to mitigate sensor noise. This filter removed measurements with fewer than $170$ neighbours in a $3$~cm radius as they were typically the result of sensor noise. The filtered measurements were then aligned to the \gls{see} pointcloud using \acrshort{icp} \citep{Besl1992}.

SEE used the same user parameters as in simulation \psecref{small-models} except for the target density, $\rho$, and the minimum separation distance, $\epsilon$ \ptblref{real-param-table}. The target density is increased by an order of magnitude and the minimum separation distance is decreased by a similar magnitude in order to capture sufficient measurements on the deer antlers after noise filtering.    

\begin{table}
	\caption{The parameters used for \gls{see} and the associated analysis in the real-world experiments on the Oxford Deer in \tsecref{real-exp}. All the values were user-specified.}
	\centering
	\begin{tabular}{@{}lll@{}}
		\toprule
		& Oxford Deer & Units \\ \midrule
		$\rho$ 				  & $5000000$ & points per m$^3$ \\
		$r$                   & $0.03$ & m \\
		$d$                   & $0.5$ & m \\
		$\epsilon$            & $0.0005$ & m \\
		$\psi$                & $0.5$ & m \\
		$\upsilon$            & $0.01$ & m \\
		$\tau$                & $100$ & number of views \\
		$\eta$                & $0.005$ & m \\
		\bottomrule
	\end{tabular}
	\tbllabel{real-param-table}
\end{table}  

The real-world performance of SEE is evaluated quantitatively and qualitatively \figref{real-results}. Its performance is quantified by surface coverage, view count, travel distance, \gls{nbv} planning time, path planning time, movement time, measurement time and total observation time. Since there is no ground truth, the surface coverage of each experiment is calculated as a percentage of the final \gls{see} pointcloud. The view count, travel distance, \gls{nbv} planning time, path planning time and movement time metrics are calculated using the methods presented in \tsecref{see-metrics}. The measurement time combines the sensor steadying time with the time required for noise filtering and applying ICP. The total observation time includes all operations from capturing the first view until SEE finishes an observation.  The qualitative performance of \gls{see} is illustrated by the final pointcloud and a mesh reconstruction generated with Open3D \citep{Zhou2018} for a representative experiment taking the median number of views (\tfigref{real-results}, top).  

The quantitative results show that SEE was as or more efficient in the real world than in simulation. The Oxford Deer required a similar number of views and movement time to the small model simulation experiments. It required less travel distance as AIT* found paths that prioritised rotating the turntable over moving the end effector by sampling more goal poses. These paths were simpler, took less planning time to find and resulted in shorter end-effector travel distances. The total observation time is greater since it measures the overall time elapsed between capturing the initial view and \gls{see} completing an observation. This includes measurement time, which was not quantified for the simulation experiments.

The qualitative results show that \gls{see} captured complete and largely accurate observations of the Oxford Deer, despite the presence of sensor noise. The coloured pointcloud (\tfigref{real-results}, top center) has high fidelity. This results in a mesh reconstruction (\tfigref{real-results}, top right)  that is also highly complete and accurate, except for some noise around the face, antlers and hind legs. These quantitative and qualitative results show that \gls{see} is suitable for real robotic scenarios. 

Further real world demonstrations of \gls{see} working with different sensor platforms (e.g., a handheld Velodyne VLP-16 LiDAR or Intel RealSense D435 camera and an Ouster OS1-64 LiDAR mounted on aerial platform) in environments with varying size and complexity (e.g., small indoor scenes to large industrial buildings) are presented in \cite{BorderThesis} and \cite{Border2023}.            

\section{Conclusion}
\seclabel{conc}

\gls{nbv} planning is key to obtaining 3D scene observations. \gls{nbv} approaches determine where sensor measurements should be captured, with the aim of efficiently obtaining a complete observation. Most existing approaches represent observations by aggregating measurements into an external scene structure. These rigid structures can be easily evaluated but often limit the fidelity of information and can be computationally expensive to maintain. This paper presents a \gls{nbv} approach that aims to overcome these limitations by using a density-based pointcloud representation.

\gls{see} is a measurement-direct \gls{nbv} approach that makes view planning decisions directly from sensor measurements to capture a minimum measurement density. Fully and partially observed surfaces are identified by individually classifying each measurement based on the density of neighbouring measurements. Measurements that lie on the boundary between these regions are classified as frontiers. Views are proposed to capture new measurements around these frontier points. These views are initially generated by considering the local surface geometry but can be refined to proactively avoid known occlusions. Observation efficiency is prioritised by choosing next best views to capture the most frontier points while moving short distances. If a view is unsuccessful then it is reactively adjusted to avoid a previously unknown occlusion or surface discontinuity. \gls{see} completes an observation when all frontier points have been observed or are deemed unobservable.

Simulation experiments comparing \gls{see} with volumetric \gls{nbv} approaches demonstrate the superior observation performance of this measurement-direct \gls{nbv} approach. \gls{see} is able to obtain highly complete observations of both small- and large-scale scene models while travelling shorter distances and requiring less observation time than the evaluated volumetric approaches. Real-world experiments conducted with a robot arm show that SEE performs equally well in the real world. \gls{see} captured high-quality observations of a deer statue using a UR10 arm with an Intel RealSense L515 sensor. Work has since demonstrated its utility on an aerial platform capable of autonomously mapping buildings \citep{Border2023}. 

An open-source implementation of \gls{see} is available at \url{https://robotic-esp.com/code/see}.          

\begin{acks}
The authors would like to thank Wayne Tubby and the Hardware Engineering team at the \gls{ori} for building the UR10 platform used for the real-world experiments.

This research was funded by UK Research and Innovation and EPSRC through Robotics and Artificial Intelligence for Nuclear (RAIN) [EP/R026084/1], ACE-OPS: From Autonomy to Cognitive assistance in Emergency OPerationS [EP/S030832/1], and the Autonomous Intelligent Machines and Systems (AIMS) Centre for Doctoral Training (CDT) [EP/S024050/1].
\end{acks}

\appendix

\section{Index to Multimedia Extensions}
The multimedia extensions in \ttblref{vid-ext} are available at \url{https://www.youtube.com/c/roboticesp}.

\begin{table}[h]
	\caption{Index to Multimedia Extensions.}
	\centering
	\resizebox{\columnwidth}{!}{%
		\begin{tabular}{l|l|p{5cm}}
			\toprule
			Extension & Media Type & Description                                                           \\ \midrule
			1         & Video      & Simulation experiments on the Newell Teapot \\
			2         & Video      & Simulation experiments on the Stanford Bunny \\
			3         & Video      & Simulation experiments on the Stanford Dragon \\
			4         & Video      & Simulation experiments on the Stanford Armadillo \\
			5         & Video      & Simulation experiments on the Happy Buddha \\
			6         & Video      & Simulation experiments on the Helix \\
			7         & Video      & Simulation experiments on the Statue of Liberty \\
			8         & Video      & Simulation experiments on the Radcliffe Camera \\
			9         & Video      & Simulation experiments on the Notre-Dame de Paris \\
			10        & Video      & Experiment on the UR10 platform \\
			\bottomrule
		\end{tabular}%
	}
	\tbllabel{vid-ext}
\end{table}

\bibliographystyle{SageH}
\bibliography{library.bib}

\end{document}